\documentclass[notitlepage,12pt]{jedm}
\usepackage[table]{xcolor}
\usepackage{url}
\usepackage{hyperref}
\usepackage{graphicx}
\usepackage{amssymb}
\usepackage{amsmath}
\usepackage{subcaption}
\hypersetup{
  colorlinks   = true,
  urlcolor     = blue,
  linkcolor    = blue,
  citecolor    = blue
}

\begin{document}

\title{Explainable AI and Machine Learning for Exam-based Student Evaluation: Causal and Predictive Analysis of Socio-academic and Economic Factors}
\author{ 
  Bushra Akter\textsuperscript{1}, 
  Md Biplob Hosen\textsuperscript{1,2,*},
  Sabbir Ahmed\textsuperscript{3},
  Mehrin Anannya\textsuperscript{1},
  Md. Farhad Hossain\textsuperscript{4}\\
  \\
  \textsuperscript{1}Institute of Information Technology, Jahangirnagar University, Dhaka, Bangladesh \\
  \textsuperscript{2}University of Maryland Baltimore County, Maryland, United States \\
  \textsuperscript{3}University of Texas at Dallas, Richardson, Texas, United States \\
  \textsuperscript{4}Cefalo, Dhaka, Bangladesh \\
  \textsuperscript{*}Corresponding author: \texttt{biplob.hosen@juniv.edu}
}

\date{}
\maketitle

\begin{abstract}
Academic performance depends on a multivariable nexus of socio-academic and financial factors. This study investigates these influences to develop effective strategies for optimizing students' CGPA. To achieve this, we reviewed various literature to identify key influencing factors and constructed a initial hypothetical causal graph based on the findings. Additionally, an online survey was conducted, where 1,050 students participated, providing comprehensive data for analysis. Rigorous data preprocessing techniques, including cleaning and visualization, ensured data quality before analysis. Causal analysis validated the relationships among variables, offering deeper insights into their direct and indirect effects on CGPA. Regression models were implemented for CGPA prediction, while classification models categorized students based on performance levels. Ridge Regression demonstrated strong predictive accuracy, achieving a Mean Absolute Error of 0.12 and a Mean Squared Error of 0.023. Random Forest outperformed in classification, attaining an F1-score near perfection and an accuracy of 98.68\%. Explainable AI techniques such as SHAP, LIME, and Interpret enhanced model interpretability, highlighting critical factors such as study hours, scholarships, parental education, and prior academic performance. The study culminated in the development of a web-based application that provides students with personalized insights, allowing them to predict academic performance, identify areas for improvement, and make informed decisions to enhance their outcomes.

{\parindent0pt
\textbf{Keywords:} Student Evaluation, Causal Analysis, Predictive Analysis, Socio-Academic Factors, Economic Factors
}

\end{abstract}

\section{Introduction}
\label{introduction}

The education system in Bangladesh, characterized by its highly competitive structure, places substantial emphasis on academic achievements, particularly the Cumulative Grade Point Average (CGPA). As the primary metric for academic performance, the CGPA serves as a key determinant of future opportunities for students, including access to advanced education, scholarships, and job placements. In Bangladesh, students are under continuous pressure to achieve a high CGPA, which not only impacts their academic reputation but also has broader implications for their personal and social lives. Failure to maintain a competitive CGPA can lead to severe consequences, such as academic probation or even dropout, which are more common than often realized ( \cite{n1,n2}). This system, while striving to maintain high standards, also exposes students to risks related to academic stress and potential burnout, with low CGPA often correlating with decreased motivation and higher dropout rates (\cite{n3}). Consequently, CGPA holds significant weight in shaping students' academic trajectories, making it an essential factor not only for students themselves but also for educators and institutions aiming to foster positive academic environments. Understanding and accurately predicting CGPA could thus support students in better managing their academic journeys, offering early interventions for those at risk, and allowing educators to tailor their approaches to student needs. However, despite the importance of CGPA in the educational landscape of Bangladesh, the underlying factors that affect students' CGPA outcomes remain insufficiently explored in a systematic manner, especially through the lens of predictive analytics and advanced causality-based models. In recognizing CGPA as a significant academic metric, this study intends to bridge this gap by identifying the complex influences on CGPA and developing predictive tools that enable a more nuanced understanding of student performance.

Despite the importance of CGPA, there exist several challenges in accurately predicting it and identifying the factors that most influence it. A primary problem in current CGPA prediction research is the lack of comprehensive factor analysis. \cite{n4} showed that while demographic and academic indicators such as age, gender, and prior academic performance are commonly studied, psychological, social, and external factors remain underexplored. \cite{n5} demonstrated that factors like motivation and self-regulation have a significant impact on academic outcomes, but these are rarely integrated into prediction models. \cite{n6} further emphasized the role of resilience and psychological well-being on students’ CGPA, indicating the need for models that consider these non-academic influences.

Another challenge lies in the reliability of prediction models. ML models have been developed to identify students at risk of academic failure, employing techniques like Artificial Neural Networks (ANN) and algorithms such as RepTree, k-NN, and Naïve Bayes (\cite{l5,l8}). These models use performance metrics like quiz scores, attendance, and academic records to predict student outcomes. The main limitation here is the reliance on a narrow set of input features, which may not capture all factors influencing student performance. Predictive models for academic achievement and student success/failure have also been developed, focusing on broader characteristics (\cite{l6,l7}). Applied to secondary school students, these models use socioeconomic, demographic, and course grade data to predict outcomes. While they aim to support proactive measures to improve passing rates and educational outcomes, they often overlook qualitative factors such as extracurricular activities or employment prospects. Some studies have introduced methods to predict CGPA and final semester grades using demographic and academic data (\cite{l9,l10}). These models leverage real-world educational datasets to enhance predictions. However, issues related to data quality and institutional variability can affect the generalizability of these models. Moreover, relying on historical academic records and specific features might overlook dynamic factors affecting student outcomes, limiting the broader applicability of these models.

Lastly, CGPA prediction models in existing literature frequently lack transparency and explainability, which poses significant challenges for both educators and students. For educators, the inability to interpret model predictions makes it difficult to identify specific factors that influence academic outcomes, limiting their capacity to design effective interventions or tailor strategies to individual student needs. For students, the lack of clear insights into how their behaviors or circumstances impact their CGPA hinders self-improvement efforts and prevents them from taking actionable steps to address areas of concern. \cite{n8} emphasized that interpretability is seen as a collaborative process that fosters trust between people and models, as well as within organizations. \cite{n9}  proposed a framework that characterizes stakeholders based on their knowledge types and contexts, helping to identify gaps in interpretability research. \cite{n7} noted that this lack of clarity limits the actionability of CGPA prediction models for decision-makers. Given these gaps, our study aims to develop a model that not only predicts CGPA accurately but also provides transparent, actionable insights to improve academic outcomes for students and stakeholders.

A critical evaluation of these approaches reveals a common limitation: the excessive focus on algorithmic improvements without sufficient consideration of real-life factors such as social, academic, and economic situations. This oversight highlights a fundamental challenge—the need to ensure that predictive models are firmly rooted in variables that have a direct, validated impact on educational outcomes, as indicated by educational psychology and related disciplines. 

To address these issues, this study proposes a novel solution that integrates machine learning and causality-based approaches to develop a transparent CGPA prediction model. Our contributions are fourfold: First, we conduct a thorough factor analysis using extensive data collection, encompassing academic, psychological, and environmental variables to ensure a holistic view of the factors influencing CGPA. Therefore, key factors from the literature that genuinely affect academic performance have been identified, as illustrated in Table ~\ref{t1}. Second, we employ state-of-the-art machine learning regression and classification models to predict CGPA with high accuracy. Third, we apply causality detection techniques, including the PC algorithm, ICA-Lingam, GES, and Grasp, to identify underlying causal relationships rather than mere correlations, offering deeper insights into the factors directly affecting CGPA outcomes. Fourth, we implement Explainable AI (XAI) methods, specifically SHAP, LIME, and interpret, to make our model’s predictions transparent and understandable, thereby empowering students and educators by providing actionable insights. Fifth, we develop a web application designed to benefit both students and teachers by providing a user-friendly interface to analyze, interpret, and act on CGPA-related insights. The primary purpose of this work is to create a reliable, explainable, and actionable model that enables students and educators to understand and act on factors influencing academic performance, potentially reducing dropout rates and improving educational strategies within the Bangladeshi context.

The structure of this work is organized as follows: First, we identify the problem and outline our objectives. Second, the methodology section describes how the work was conducted, including data collection, factor analysis, machine learning models, application of causality techniques, XAI methods, and the development of the web application. Third, the results section presents the findings, including model performance, causality insights, XAI-driven interpretations, and the web application’s features. At last, the conclusion highlights the implications of the work, its contributions to educational analytics, and future directions.

\begin{table}
  \caption{Factors Influencing Student Performance and Their Acronyms with References. Each factor is categorized by type and effect, with reference citations where applicable.}
  \Description{A six-column table listing factors affecting student performance, including their acronyms and summarized effects, with cited references. For example, S.S.C Result (GPA) has acronym SSC and effect 'Motivation', referenced by lt4.}
  \label{t1}
  \centering
  \renewcommand{\arraystretch}{1.5}
  \begin{tabular}{|p{2.5cm}|p{1.3cm}|p{4cm}|p{2.5cm}|p{1.3cm}|p{4cm}|}
    \hline
    \textbf{Factors} & \textbf{Acron ym} & \textbf{Effect/Note [Ref]} & \textbf{Factors} & \textbf{Acron ym} & \textbf{Effect/Note [Ref]} \\
    \hline
    Department/ Institute & DI & Less effect & S.S.C Result (GPA) & SSC & Motivation \cite{lt4} \\
    \hline
    Year/Semester & YS & No effect & H.S.C Result (GPA) & HSC & Motivation and confidence \cite{lt4} \\
    \hline
    Gender & G & Sometimes influential \cite{lt5,lt6} & Father Education (In Year) & FE & Influential \cite{lt5,lt6} \\
    \hline
    Father Job & FJ & Influential \cite{lt5,lt6} & Mother Education & ME & Influential \cite{lt5,lt6} \\
    \hline
    Mother Job & MJ & Influential \cite{lt5,lt6} & Major Illness & MI & Impairing \cite{lt7} \\
    \hline
    Attendance in Class & AC & Consistency \cite{lt9} & Group Study & GS & Collaborative \cite{lt18,lt19} \\
    \hline
    Study Hour (in a week) & SH & Discipline & Political Involvement & PI & Distracting \cite{lt17} \\
    \hline
    Sport/Cultural Involvement & S & Enriching \cite{lt20,lt21,lt22} & Internet Facilities & IF & Distracting and resources \cite{lt8,lt14,lt15,lt16} \\
    \hline
    Getting Any Scholarship & PSR & Confidence \cite{lt10,lt11} & Hostel Staying & HS & Independence \cite{lt23,lt24,lt25} \\
    \hline
    Self-Income (in taka) & C & Confidence \cite{lt27} & Relational Status & RS & Distracting and supportive \cite{lt12,lt13} \\
    \hline
    Communication Skill & CS & Influence \cite{lt26} & Confidence & SCI & Correlation \cite{lt27} \\
    \hline
  \end{tabular}
\end{table}

\section{Materials and Methods} 

\begin{figure}
  \centering
  \includegraphics[width=0.95\textwidth]{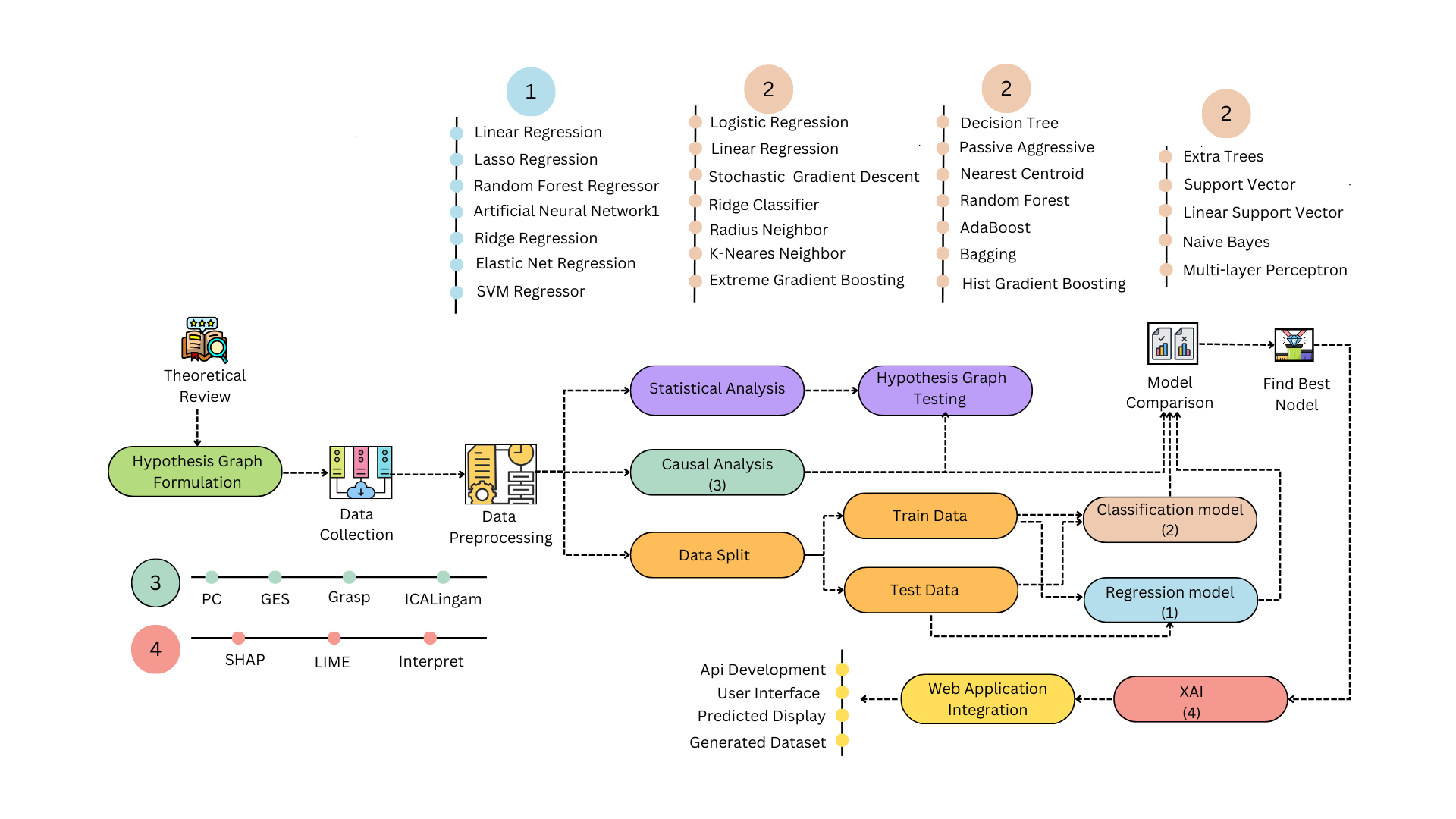}
  \Description{A diagram showing the overall workflow of the study including input data, causal inference, predictive modeling, XAI techniques, and final evaluation stages.}
  \caption{Overall workflow diagram of the study.}
  \label{workflow}
\end{figure}

This work procedure, as illustrated in Fig~\ref{workflow}, begins with the formulation of an initial hypothesis graph, serving as the foundation for analyzing causal relationships between variables. The workflow includes stages such as data collection, preprocessing, and cleaning, with the preprocessing phase further divided into statistical analysis, causal analysis, and data splitting for training and testing. Hypothesis graph clustering is employed during statistical and causal analysis to uncover relationships among features, while regression and classification techniques are applied after the data split. Following model development and evaluation, XAI techniques, including SHAP, LIME, and Interpreter, are employed to enhance transparency and interpretability.

\begin{figure}
  \centering
  \includegraphics[width=0.95\textwidth]{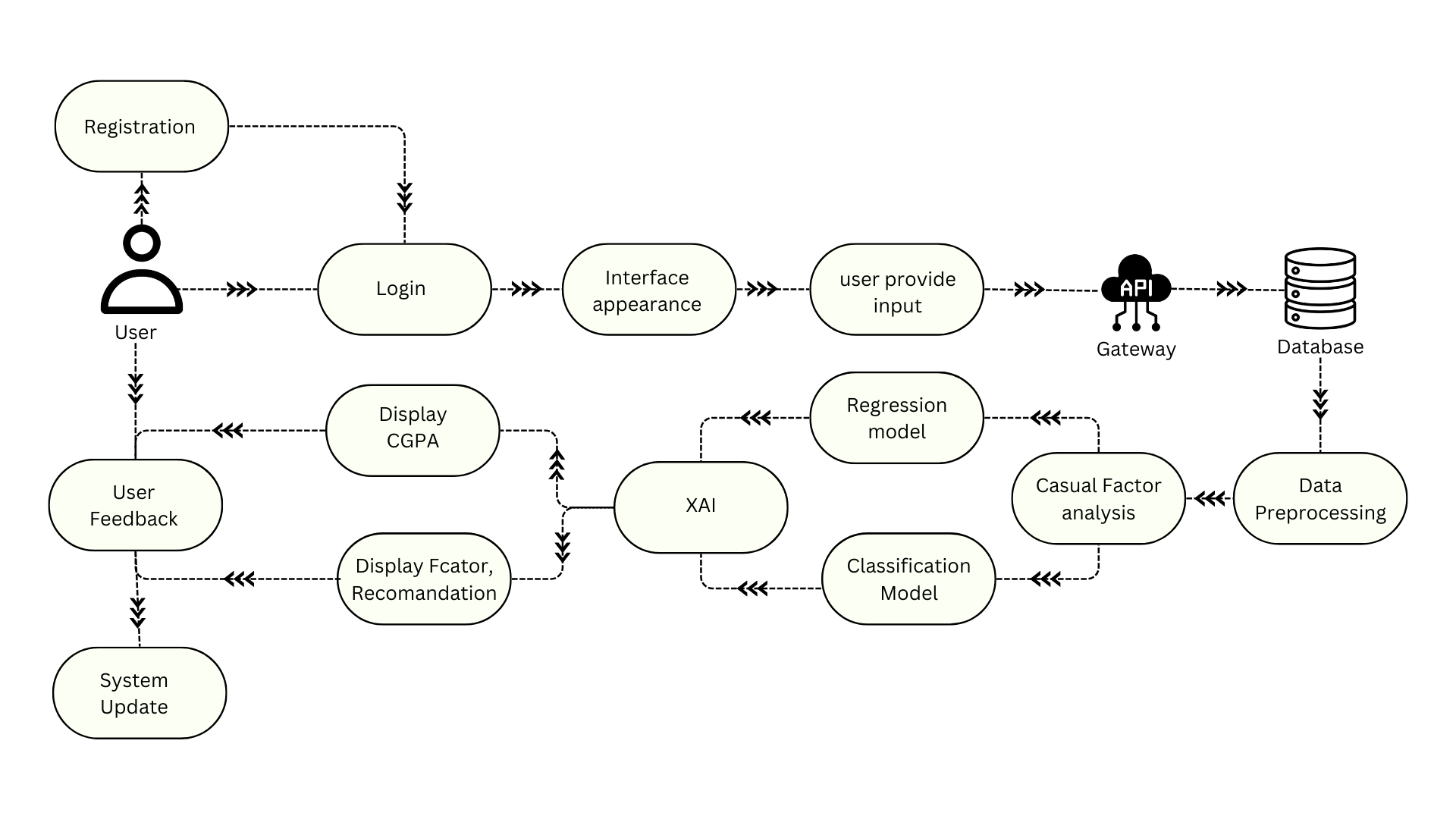}
  \Description{A service flow diagram illustrating the interaction within the CGPA prediction system, starting from user input, model prediction, feedback integration, and explanation output.}
  \caption{User service flow diagram depicting the step-by-step interaction within the CGPA prediction system, from user input to feedback integration.}
  \label{wo}
\end{figure}

At the final stage, the prediction model is deployed in a web application, allowing users to interact with the system seamlessly. As illustrated in Fig~\ref{wo}, the web application guides users step by step, enabling them to register, input academic and socio-economic data, and receive CGPA predictions along with an analysis of influencing factors. The system not only predicts CGPA but also provides personalized recommendations to help students improve their performance. Users can further give feedback on the predictions, which is stored and used to refine the model iteratively. This dynamic feedback loop ensures continuous enhancement of prediction accuracy, making the system a robust and adaptive academic performance analysis tool.

\subsection{Hypothesis Causal Graph}

\begin{figure}[!ht]
  \centering
  \includegraphics[width=0.95\textwidth]{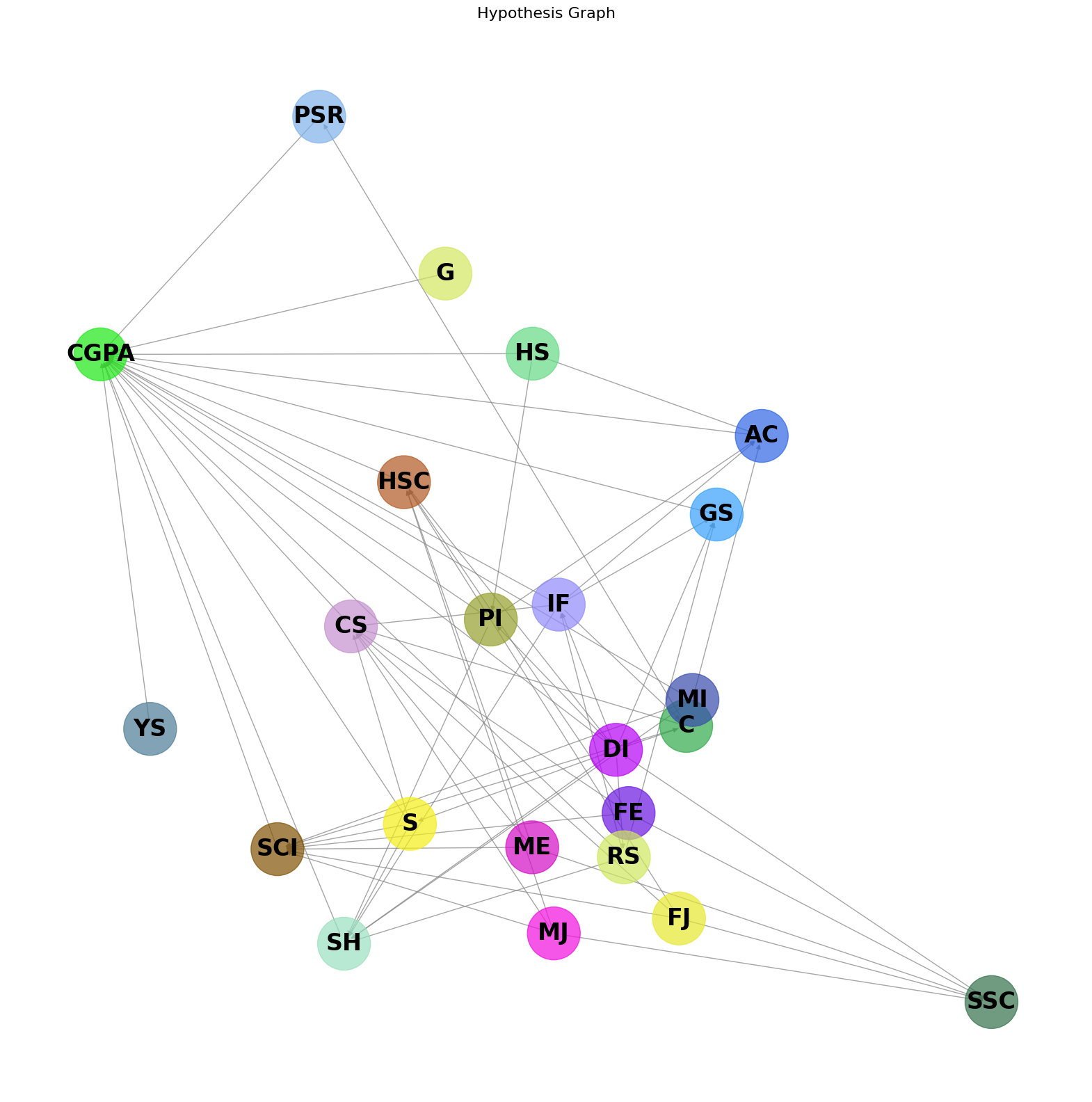}
  \Description{A hypothesis-level causal graph showing the relationships among variables such as socioeconomic status, attendance, study hours, and CGPA based on literature review.}
  \caption{Hypothesis causal graph illustrating the relationships among variables influencing CGPA, based on literature insights.}
  \label{causal}
\end{figure}

At first, we draw a hypothesis graph to analyze key factors influencing CGPA, derived from literature and real-world scenarios. Insights from the literature, summarized in Table \ref{t1}, highlight foundational academic contributors such as HSC and SSC grades, which influence CGPA directly and indirectly through DI. Personal factors like ME, FE, MJ, and FJ affect CGPA indirectly by shaping intermediate variables. These intermediates, including S, RS, and PI, act as pathways through which foundational and personal factors influence CGPA. S is identified as a critical factor for academic success, while RS highlights the role of external support. PI serves as a mediator, linking DI and RS to CGPA. Using these insights, the hypothesis causal graph in Fig~\ref{causal} illustrates the intricate relationships among these variables and provides a structured framework for analyzing the impact of each factor.

\subsection{Data Collection}

Based on the causal graph of the hypothesis, a dataset of 1,050 university-going student profiles was collected, incorporating all variables necessary to examine the factors influencing CGPA. Academic indicators such as DI, YS, SSC, HSC, and CGPA were used to represent foundational contributors, while familial and demographic factors like FE, ME, FJ, and MJ were also included. Personal attributes (MI, AC, SH, IF, S, HS), socioeconomic factors (PSR, C, RS), and skill-based measures (CS, SCI) were collected to ensure comprehensive coverage of the hypothesis. All participants were university students aged 18 years or older, complying with the standard legal age of majority. Data were collected via a Google Form, where electronic informed consent was obtained through a mandatory checkbox at the beginning of the form. This checkbox confirmed that each participant voluntarily agreed to take part in the study, acknowledged understanding of the study’s purpose, and consented to the anonymous use of their data for academic research. No personally identifiable information was collected, and all data were handled with strict confidentiality. The study received ethical approval from the Biosafety, Biosecurity and Ethical Committee, Faculty of Biological Sciences, Jahangirnagar University, Savar, Dhaka-1342, on 17 February 2025, under reference number BBEC, JUM 2025/02 (207). Detailed characteristics of the dataset, including unique values and any missing entries, are presented in Table \ref{tab2}, ensuring transparency and data integrity.

\subsection{Data Cleaning, Processing \& Statistical Analysis }

The dataset information in Table \ref{tab2} reveals that there are no null values, indicating it is complete data. We removed duplicate entries to ensure data integrity.

For categorical variables such as G, FJ, MJ, MI, IF, CS, SCI, HS, SH, RS, GS, and PI, encoding techniques were applied to convert them into numerical representations, enabling their inclusion in the analysis. Continuous variables such as SSC, HSC, and CGPA were standardized to ensure uniformity in scale. After completing these preprocessing steps, the dataset was confirmed to be ready for analysis.

Statistical analysis then focused on understanding the relationships between various factors. First, the direct impact of SH, GS, AC, RS, FJ, MJ, SCI, and CS on CGPA was explored using correlation analysis and visual plots. These analyses highlighted which factors significantly influenced academic outcomes. Additionally, relationships among behavioral variables were examined. For instance, the connections between SH, HS, GS, S, SCI, C, AC and G were analyzed to identify patterns and trends. This step aimed to uncover how external and personal factors interact to shape study habits and academic behavior.

\begin{table*}[!htbp]
  \centering
  \caption{Descriptive statistics of dataset features.}
  \label{tab2}
  \begin{tabular}{p{1.5cm}p{1.6cm}p{1.5cm}p{2.9cm} | p{1.5cm}p{1.6cm}p{1.5cm}p{2.9cm}}
    \hline
    Factors & Non-null count & Unique values & Most frequent value & 
    Factors & Non-null count & Unique values & Most frequent value \\
    \hline
    DI & 1050 & 63 & IIT & YS & 1050 & 10 & 3.0 \\
    G & 1050 & 2 & Female & SSC & 1050 & 38 & 5 \\
    HSC & 1050 & 47 & 5 & FE & 1050 & 26 & 16 \\
    FJ & 1050 & 7 & Govt. job & ME & 1050 & 26 & 10 \\
    MJ & 1050 & 5 & Unemployed & MI & 1050 & 2 & No \\
    AC & 1050 & 3 & 75–100\% & SH & 1050 & 4 & 3–9 hours \\
    IF & 1050 & 3 & Available & GS & 1050 & 3 & Participate \\
    S & 1050 & 2 & No & PI & 1050 & 2 & No \\
    HS & 1050 & 3 & Regular & PSR & 1050 & 2 & No \\
    C & 1050 & 4 & 3000–5000 & RS & 1050 & 3 & Single \\
    CS & 1050 & 3 & Average & SCI & 1050 & 23 & Not enough confident \\
    CGPA & 1050 & 114 & 3.63 & & & & \\
    \hline
  \end{tabular}
\end{table*}

\subsection{Prediction By Regression }

We based our approach on applying multiple regression models to identify the one that provides the most accurate CGPA predictions. By evaluating the performance of each model, we aimed to determine which technique best captures the underlying relationships in the dataset. The comparison was conducted using metrics like MAE, MSE, and RMSE, ensuring a robust assessment of each model's predictive capabilities.

At first, we applied Linear Regression as the baseline model, offering simplicity and straightforward interpretation. It highlighted linear relationships between predictors and CGPA, providing a foundation for comparison. To address the influence of outliers, Ridge Regression was applied, leveraging L2 regularization to shrink coefficients and reduce overfitting. Lasso Regression, with its L1 regularization, further refined predictions by effectively performing feature selection, setting insignificant coefficients to zero. Elastic Net Regression, which combines L1 and L2 penalties, provided a balanced approach, making it particularly effective for datasets with numerous interrelated predictors.

Moving beyond linear methods, Random Forest Regressor was utilized as an ensemble technique, combining predictions from multiple decision trees to capture non-linear relationships and improve accuracy. The SVM Regressor provided robust predictions by maximizing the margin around the hyperplane, excelling in handling outliers and complex data patterns. Finally, ANN explored deeper, non-linear relationships through multi-layer architectures, experimenting with different configurations to optimize performance.

\subsection{Prediction By Classification}

For predicting students' CGPA, we used various classification algorithms. The Logistic Regression classifier handled binary classification tasks by modeling probabilities using the logistic function. The Ridge Classifier combined linear regression with L2 regularization for improved classification performance. The Decision Tree classifier created models by splitting data based on feature values, providing interpretability. The K-Nearest Neighbor (KNN) classifier assigned class labels based on the closest data points using proximity measures.

The Extreme Gradient Boosting (XGB) classifier, a powerful and scalable implementation of gradient boosting, enhanced predictive performance through iterative refinement. The Random Forest  classifier, an ensemble method utilizing multiple decision trees, improved accuracy and reduced overfitting. The Hist Gradient Boosting classifier utilized histograms for faster computation, making it suitable for large datasets. The Support Vector Machine (SVM) classifier maximized the margin between data points and the decision boundary, reducing overfitting. The Multi-layer Perceptron (MLP) classifier, leveraging neural networks, captured intricate relationships in the data for complex pattern recognition. The AdaBoost  classifier combined weak classifiers to form a strong predictive model, enhancing classification performance.

\subsection{Unsupervised Causality Analysis}

The identification of causal relationships among the variables was performed using four advanced algorithms: the PC algorithm, GES, GRaSP, and ICALingam. These algorithms rely on different approaches to model and infer causal structures, with the relationships being represented as directed acyclic graphs (DAGs). 

\subsubsection*{PC Algorithm}

The PC algorithm utilizes a principle based on conditional independence to determine causal relationships. The approach begins by assuming that all variables are potentially causally related, and edges are iteratively removed based on conditional independence tests. The fundamental hypothesis that drives this process is expressed in Equation \ref{pc}, where it is hypothesized that if two variables, \( X \) and \( Y \), are conditionally independent given a set of variables \( Z \), the edge between them is removed. This is represented as:

\begin{equation}
\label{pc}
    P(X \perp Y \mid Z) \implies \text{Remove the edge between } X \text{ and } Y.
\end{equation}

The process begins by testing conditional independence between each pair of variables, adjusting the graph structure by removing edges when conditional independence is detected. This iterative procedure continues, refining the graph structure by progressively eliminating relationships that are statistically insignificant, thus revealing the underlying causal dependencies among the variables.

\subsubsection*{GES Algorithm}

The GES algorithm approaches causality from a score-based perspective, specifically focusing on maximizing the Bayesian Information Criterion (BIC). The BIC score helps determine the model that best balances goodness-of-fit with model complexity. The BIC formula, as shown in Equation \ref{gis}, is:

\begin{equation}
\label{gis}
    \text{BIC} = \log L(\mathcal{M}) - \frac{k}{2} \log N,
\end{equation}

where \( L(\mathcal{M}) \) is the likelihood of model \( \mathcal{M} \), \( k \) represents the number of free parameters, and \( N \) is the sample size. GES operates by first searching for an initial graph structure, then iteratively adding and removing edges to optimize the BIC score. In the first phase, the algorithm begins with an empty graph and adds edges that maximize the BIC. In the second phase, edges are removed if their removal increases the BIC, resulting in the most statistically optimal model. The final graph that emerges from this process represents the causal structure that best explains the data while avoiding unnecessary complexity.

\subsubsection*{GRaSP Algorithm}

GRaSP builds on the GES algorithm but introduces a regularization term that encourages sparsity in the graph. This additional constraint prevents the model from overfitting to the data by penalizing the inclusion of too many edges. The loss function used in GRaSP is shown in Equation \ref{grasp}:

\begin{equation}
\label{grasp}
    \mathcal{L}(\mathcal{G}) = \text{Score}(\mathcal{G}) - \lambda ||\mathcal{G}||_1,
\end{equation}

where \( \lambda \) is a regularization parameter that controls the degree of sparsity, and \( ||\mathcal{G}||_1 \) is the \( l_1 \)-norm of the adjacency matrix of the graph \( \mathcal{G} \). The term \( ||\mathcal{G}||_1 \) penalizes the model for adding more edges, thereby enforcing a simpler graph structure. The objective of the GRaSP algorithm is to find a causal graph that not only explains the data well but also avoids overcomplicating the model by limiting the number of edges. This balance is achieved by optimizing both the score and the sparsity penalty.

\subsubsection*{ICALingam Algorithm}

ICALingam is based on independent component analysis (ICA) and is designed to identify causal relationships by modeling the observed data with a linear non-Gaussian structural equation model (SEM). The algorithm assumes that the observed variables \( X = (x_1, x_2, \dots, x_n) \) are generated from a linear process, and the relationship between these variables is described by the SEM in Equation \ref{ica}:

\begin{equation}
\label{ica}
    X = B X + E,
\end{equation}

where \( B \) is the weighted adjacency matrix representing the causal relationships, and \( E \) is the vector of independent noise terms. ICA is used to separate the independent components of the observed data, enabling the identification of the underlying causal structure. To solve for the matrix \( B \), ICA minimizes the reconstruction error, as shown in Equation \ref{ica2}:

\begin{equation}
\label{ica2}
    B = \text{argmin}_{B} \, || X - B X ||^2.
\end{equation}

The objective is to find the matrix \( B \) that best explains the observed data while ensuring that the graph remains acyclic, i.e., no feedback loops are present. ICA assumes that the noise terms are independent and non-Gaussian, which aids in identifying the underlying structure of the causal relationships. The solution to this optimization problem provides a model that reflects the true causal influences among the observed variables.

\subsection{XAI}

In this study, XAI techniques are applied to provide transparency and interpretability to the machine learning models predicting annual increments for garment sewing machine operators. Specifically, the techniques SHAP, LIME, and Interpreter are used to ensure that the models not only make accurate predictions but also provide understandable insights into the factors influencing those predictions.

\subsubsection*{SHAP}

SHAP values are used to explain the contribution of each feature to the model's prediction. By calculating the Shapley value for each feature, SHAP provides a quantitative measure of how each feature affects the final prediction. The Shapley value for a given feature \( f_j \) is calculated using the equation in \ref{shape}:

\begin{eqnarray}
\label{shape}
    \phi_j &=& \sum_{S \subseteq N \setminus \{j\}} \frac{|S|!(|N|-|S|-1)!}{|N|!} \left[ f(S \cup \{j\}) - f(S) \right]
\end{eqnarray}

where \( \phi_j \) is the Shapley value for feature \( f_j \), \( N \) is the set of all features, and \( S \) is a subset of features excluding \( j \). This equation calculates the average marginal contribution of feature \( j \) across all possible feature combinations, providing a clear understanding of its influence on the model.

\subsubsection*{LIME}

LIME is used to explain individual predictions by approximating the model locally around a specific data point. For each prediction \( f(x) \), LIME generates a local surrogate model \( g(x) \) by sampling the feature space around the instance \( x \) and training a simple interpretable model on these perturbed samples. The objective is to approximate the black-box model in the vicinity of \( x \). The local model is learned by solving the following optimization problem, as shown in Equation \ref{limee}:

\begin{eqnarray}
\label{limee}
    \min_g \sum_{i} \left[ \text{loss}(f(x_i), g(x_i)) \right] + \lambda \|g\|_1
\end{eqnarray}

where \( \text{loss} \) is the difference between the black-box model prediction and the surrogate model, and \( \|g\|_1 \) is a regularization term to keep the model interpretable. This approach helps us gain insight into how specific features, such as attendance or skills, affect the predictions for a particular operator.

\subsubsection*{Interpreter}

Interpreter provides both local and global explanations by analyzing the model’s behavior across all predictions. It generates a set of rules that describe the relationships between features and predictions. For local explanations, Interpreter approximates the model's decision boundaries for each data point, while for global explanations, it analyzes feature importance and interactions across the entire dataset. The method uses decision trees to approximate the model and visualize feature contributions. This is achieved by minimizing the following objective function for feature interaction and importance, as shown in Equation \ref{eq:interpreter}:

\begin{eqnarray}
\label{eq:interpreter}
    \min_{\theta} \sum_{i=1}^n \left[ \text{loss}(f(x_i), T(x_i, \theta)) \right]
\end{eqnarray}

where \( f(x_i) \) is the model's prediction for instance \( x_i \), and \( T(x_i, \theta) \) is the decision tree surrogate for the model, with \( \theta \) representing the tree parameters.

\subsection{Web Application Integration with the Best Predictor Model}

The final phase of this project involved the development and deployment of a web application that predicts students' CGPA using the best-performing model. After evaluating various algorithms, the optimal predictor was deployed as a microservice, enabling seamless communication via RESTful API endpoints. This architecture allowed for a smooth workflow in which user inputs are captured by the frontend, transmitted to the backend, and processed by the deployed model to generate accurate predictions.

The backend, developed using Node.js and Express.js, manages API requests and facilitates communication between the ReactJS-based frontend and the prediction model. Upon submission, user inputs are validated and processed by the backend before being forwarded to the prediction model. The model then computes the predicted CGPA and returns the result, which is stored in the predictions table within a MySQL database. The system architecture leverages a diverse set of technologies to ensure robust performance and ease of integration, including:

\begin{itemize}
    \item \textbf{Front End:} ReactJS, HTML, CSS 
    \item \textbf{Back End:} Node.js, Express.js, JWT 
    \item \textbf{Database:} MySQL, Sequelize, XAMPP 
\end{itemize}

In addition to its predictive capabilities, the web application incorporates causal analysis derived from the hypothetical causal graph and Explainable AI (XAI) methodologies. By examining the relationships between academic and socioeconomic factors and their impact on CGPA, the system identifies the most influential predictors, such as study hours, scholarships, parental education, and previous results. These insights are then transformed into personalized, actionable recommendations for each user. Continuous feedback from users allows these recommendations to be refined over time, ensuring that the system remains dynamic and responsive to evolving academic contexts.

\section{Experimental Results}

\subsection{Statistical Analysis}

\begin{figure}[h!]
    \centering
    \begin{minipage}{0.48\textwidth}
        \centering
        \includegraphics[width=\textwidth]{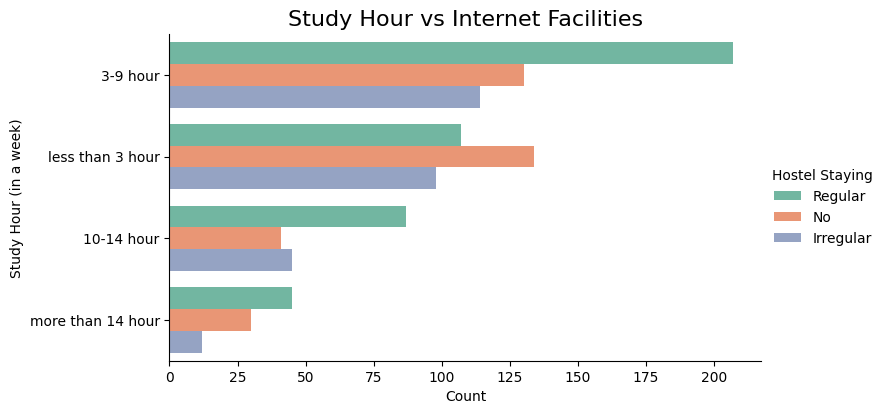}
        \subcaption{HS vs. SH}
        \label{hostel_study}
    \end{minipage}
    \hfill
    \begin{minipage}{0.48\textwidth}
        \centering
        \includegraphics[width=\textwidth]{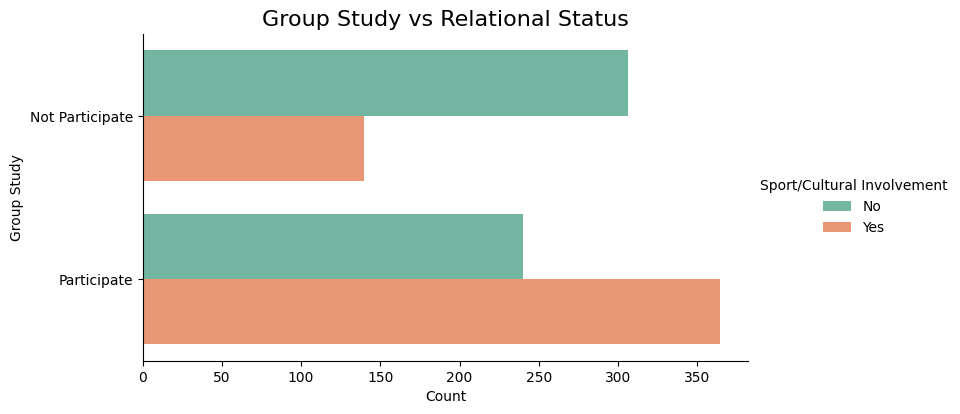}
        \subcaption{GS vs. S}
        \label{group_study_sports}
    \end{minipage}

    \vspace{0.5\baselineskip}

    \begin{minipage}{0.48\textwidth}
        \centering
        \includegraphics[width=\textwidth]{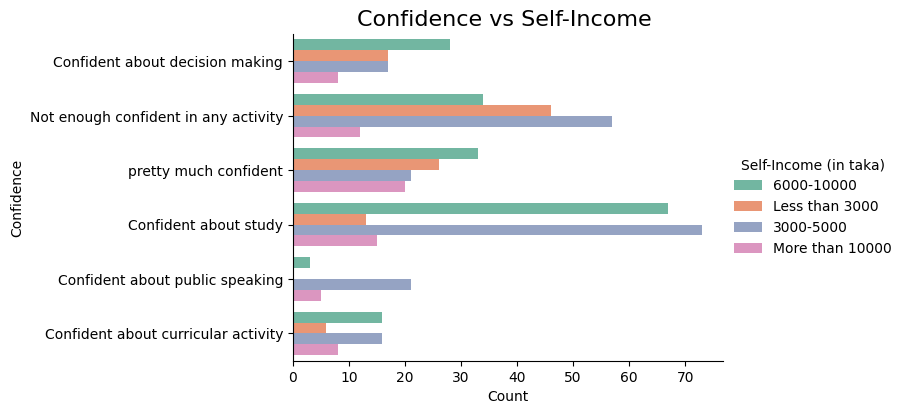}
        \subcaption{SCI vs. C}
        \label{confidence_income}
    \end{minipage}
    \hfill
    \begin{minipage}{0.48\textwidth}
        \centering
        \includegraphics[width=\textwidth]{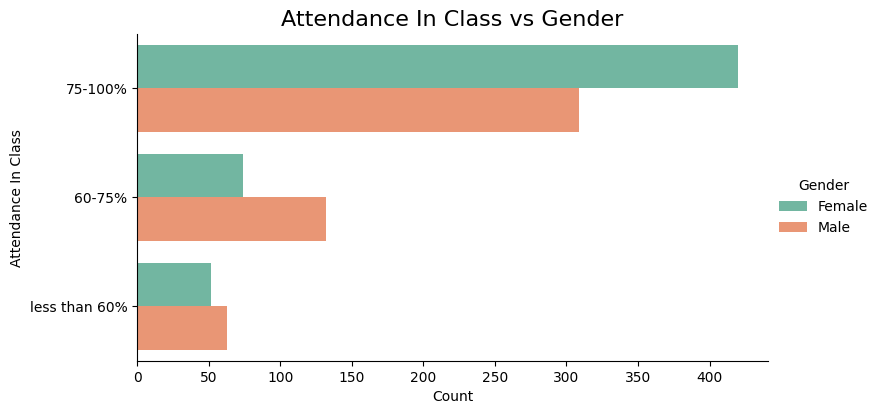}
        \subcaption{AC vs. G}
        \label{attendance_gender}
    \end{minipage}
    
    \Description{Four subfigures analyzing relationships among variables such as hostel staying vs study hour, group study vs sports, confidence vs income, and attendance vs gender.}
    \caption{Analysis of various relationships: SH, HS, GS, S, SCI, C, AC, and G, highlighting their interconnected dynamics and mutual influences.}
    \label{variable_analysis}
\end{figure}
We begin the statistical analysis by exploring the relationships between various factors such as HS, GS, C, and AC, which influence students' academic behavior and performance.

We explore the relationship between HS and SH. Students who stay in Hostel and study irregularly for 3-9 hours a week make up 25.28\% of the total, whereas those who study more than 14 hours a week represent 13.79\%, as shown in Figure~\ref{variable_analysis}(a). On the other hand, students who study more than 14 hours but do not stay in Hostel constitute 34.48\% of that group. The findings show that students in Hostel tend to study less (3-9 hours), while non-Hostel staying students study longer hours on average. This suggests that living in Hostel may introduce more distractions or less ideal study environments, leading to reduced SH. Conversely, non-HS students likely have more favorable environments for study, resulting in longer study periods.

We highlight the relationship between GS participation and S. Among students who do not participate in GS, 68.61\% are not involved in S, while 39.74\% of students who participate in GS are also engaged in S. This suggests that students who participate in GS tend to be more involved in extracurricular activities, demonstrating the potential benefits of collaborative learning in encouraging well-rounded student engagement beyond academics, as shown in Figure~\ref{variable_analysis}(b).

In Figure~\ref{variable_analysis}(c), we analyze SCI levels in various activities against C. The data indicates a clear positive relationship between C and SCI in decision-making, public speaking, and curricular activities. Higher-income students (above 10,000 BDT) display the most SCI, while students earning less than 3,000 BDT show the least SCI across all measured areas. This observation reveals the important role that financial stability plays in boosting students' self-assurance, which in turn can enhance their performance in academic and extracurricular pursuits.

We examine the effect of AC rates and G on class participation. Female students have higher AC rates, with 57.61\% attending 75-100\% of the time, while male students show a split of 42.39\% attending the same range, as shown in Figure~\ref{variable_analysis}(d). This suggests a significant difference in class attendance based on G, with female students attending more consistently. This trend emphasizes the potential influence of G on academic engagement and class participation.

\begin{figure}[!h]
    \centering
    \begin{minipage}{0.48\textwidth}
        \centering
        \includegraphics[width=\textwidth]{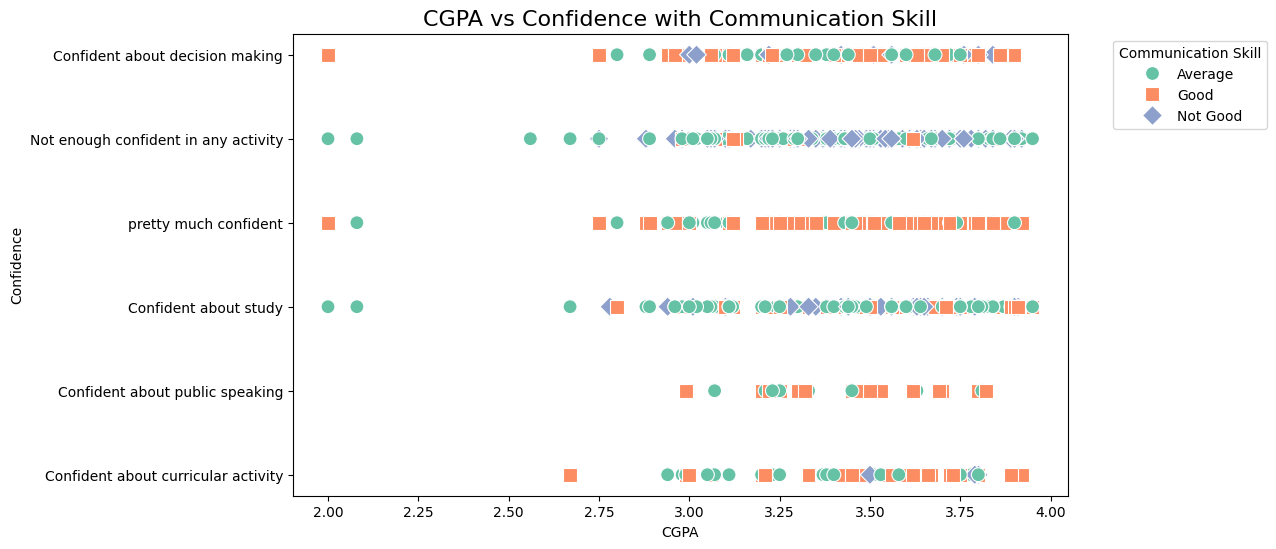}
        \subcaption{SCI and CS in relation to CGPA}
        \label{s1}
    \end{minipage}
    \hfill
    \begin{minipage}{0.48\textwidth}
        \centering
        \includegraphics[width=\textwidth]{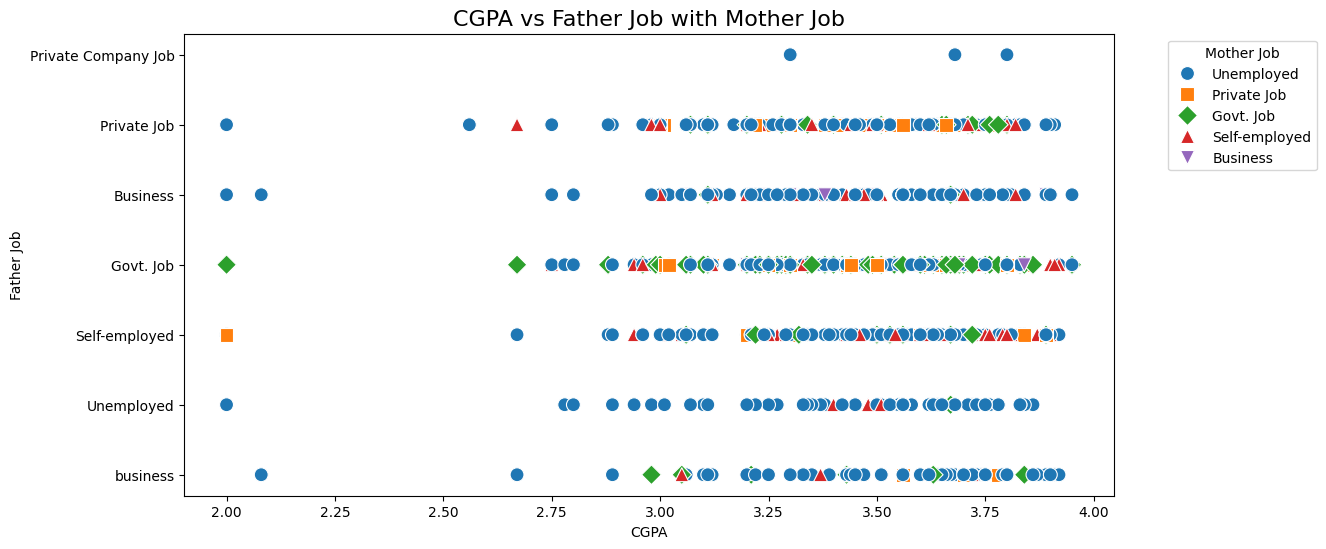}
        \subcaption{Impact of parental occupation on CGPA}
        \label{s2}
    \end{minipage}

    \vspace{0.5\baselineskip}

    \begin{minipage}{0.48\textwidth}
        \centering
        \includegraphics[width=\textwidth]{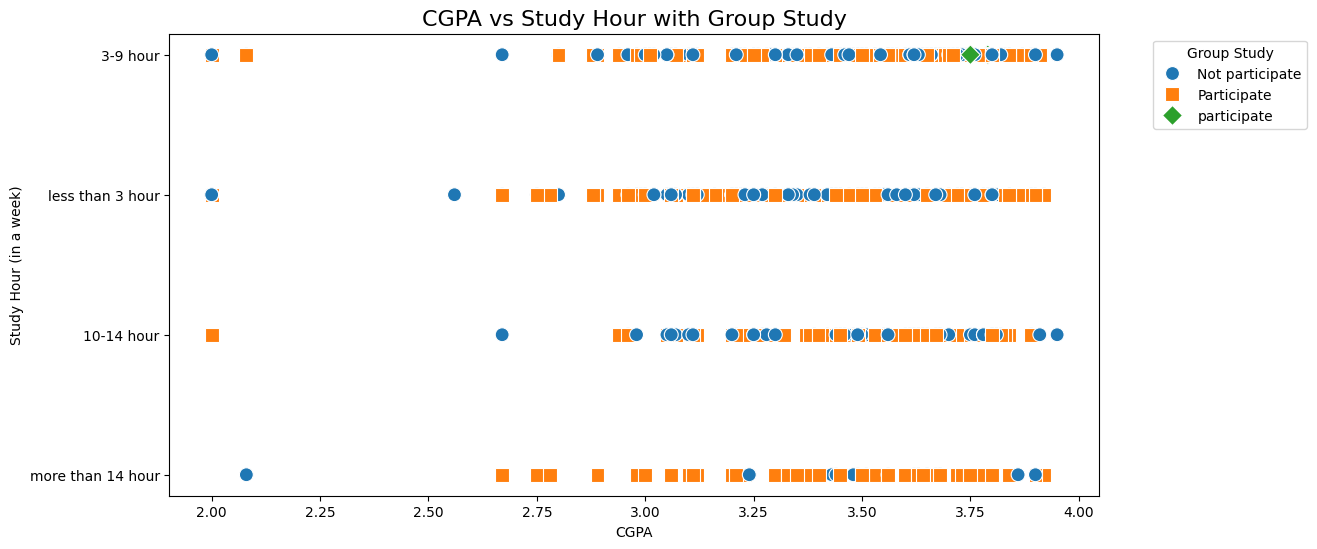}
        \subcaption{SH and GS participation influencing CGPA}
        \label{s3}
    \end{minipage}
    \hfill
    \begin{minipage}{0.48\textwidth}
        \centering
        \includegraphics[width=\textwidth]{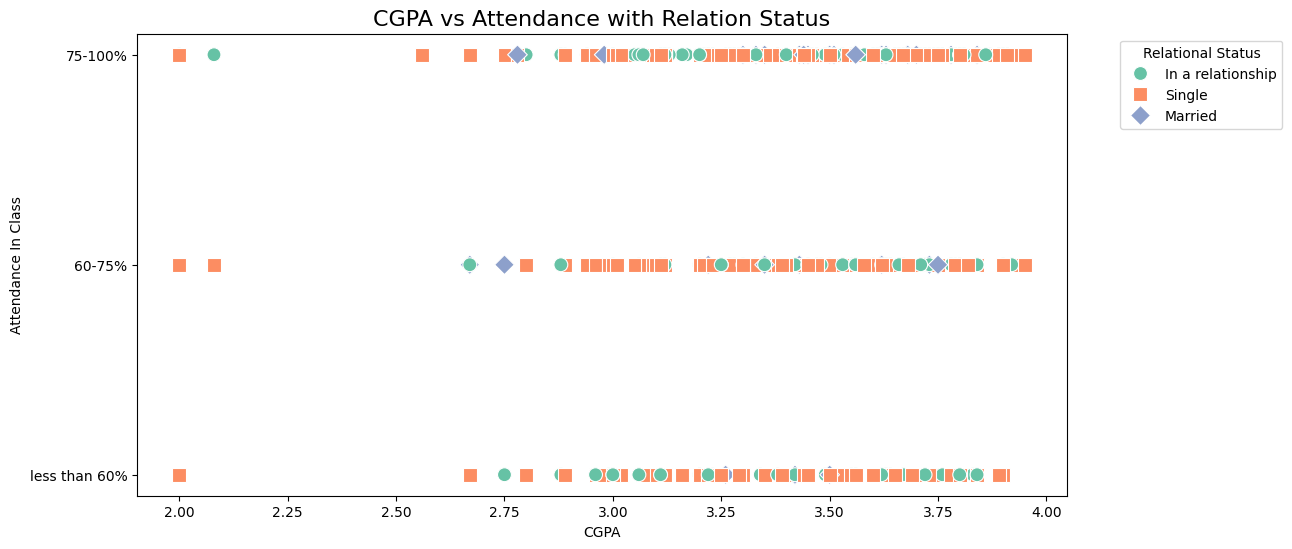}
        \subcaption{AC and RS affecting CGPA}
        \label{s4}
    \end{minipage}
    
    \Description{Four subfigures analyzing the relationship between factors such as study confidence and communication skills (SCI and CS), parental occupation, study hours and group study participation (SH and GS), attendance and relational status (AC and RS) with CGPA.}
    \caption{Analysis of the relationship between factors such as SCI, CS, parental occupation, SH, GS, AC, and RS with CGPA.}
    \label{S}
\end{figure}
After that, we analyze various factors that influence academic behavior, performance, and their impact on CGPA prediction. The analysis of the graphs in Figure~\ref{S} reveals the impact of various factors on CGPA, offering valuable insights into the complex relationships between personal attributes, study habits, and academic performance. Figure~\ref{S}(a) shows that students with higher CGPAs (3.5-4.0) tend to have better CS and SCI in curricular activities. Specifically, those with good CS consistently outperform in academics. Figure~\ref{S}(b) discusses the impact of parental occupation on CGPA. Students whose parents have stable jobs, such as government or private sector positions, tend to have higher CGPAs compared to those with self-employed or unemployed parents.In Figure~\ref{S}(c), the relationship between SH and GS on CGPA is explored. Students who study more hours (10-14 per week) and participate in GS are most likely to have higher CGPAs. Conversely, students who study less and avoid GS tend to have lower CGPAs.

Figure~\ref{S}(d) investigates how AC and RS affect CGPA. Students with high AC rates 75-100\% generally show better academic performance, with many of them falling in the 3.5-4.0 CGPA range. Interestingly, while RS shows some variation, the most consistent pattern is seen in those with higher AC. Overall, the analysis suggests that attributes such as effective CS, stable parental occupation, SH, GS involvement are key drivers of academic success, directly influencing CGPA.

\subsection{Regression Result}
\begin{figure}[!h]
  \centering
  \includegraphics[width=0.95\textwidth]{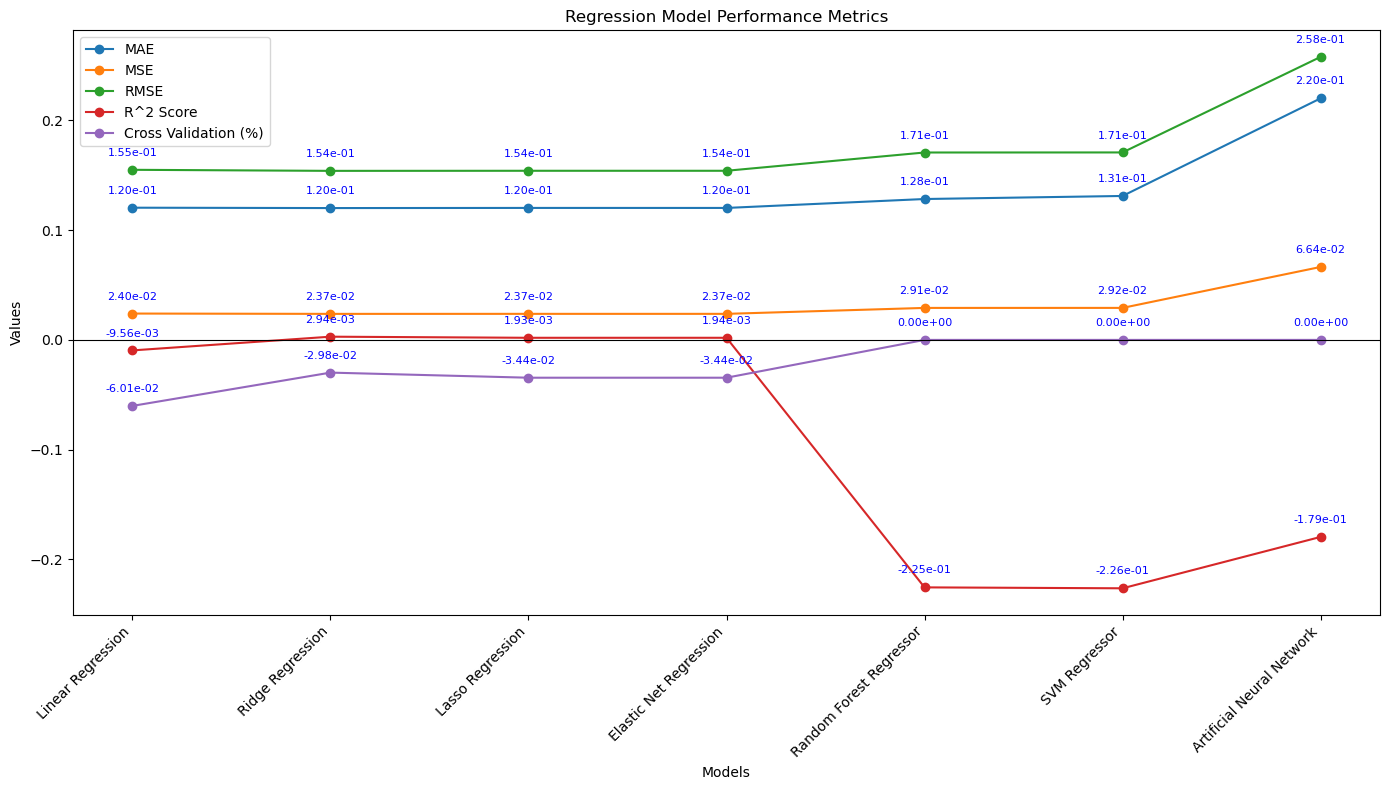}
  \Description{Bar charts comparing regression model performances for CGPA prediction, showing metrics such as MAE, MSE, RMSE, R squared, and cross-validation percentage for various models.}
  \caption{Comparison of regression model performances for CGPA prediction, evaluating metrics such as MAE, MSE, RMSE, R², and cross validation (\%) across various models.}
  \label{rr}
\end{figure}

The regression analysis results, shown in Figure \ref{rr}, reveal that Ridge Regression yielded the best performance among all the models. It achieved an MAE of 0.120, an MSE of 0.0237, and an RMSE of 0.1539, indicating relatively low error compared to the other models. While its R-squared value of 0.0029 suggested minimal improvement over the baseline Linear Regression, it still showed the best balance between predictive accuracy and generalization. In comparison, Linear Regression produced similar results with an MAE of 0.120 and an MSE of 0.024, but it had a slightly worse R-squared value of -0.0095, indicating a poor fit to the data. The Lasso Regression and Elastic Net Regression models demonstrated nearly identical performance to Ridge Regression, showing no significant benefit in terms of predictive power or generalization. On the other hand, more complex models like Random Forest Regressor, SVM Regressor, and Artificial Neural Network performed worse, with higher MAEs, MSEs, and RMSEs, along with negative R-squared values indicating poor model fit. The Random Forest Regressor showed an MAE of 0.128 and an MSE of 0.0291, while the SVM Regressor exhibited an MAE of 0.131 and an MSE of 0.0292, both struggling with generalization. The Artificial Neural Network had the worst performance with an MAE of 0.220, an MSE of 0.0664, and an RMSE of 0.2577, making it the least suitable model. Overall, Ridge Regression emerged as the most reliable model in this analysis, offering the best trade-off between accuracy and generalization.

\subsection{Classification Result}

\begin{figure}[!h]
  \centering
  \includegraphics[width=0.95\textwidth]{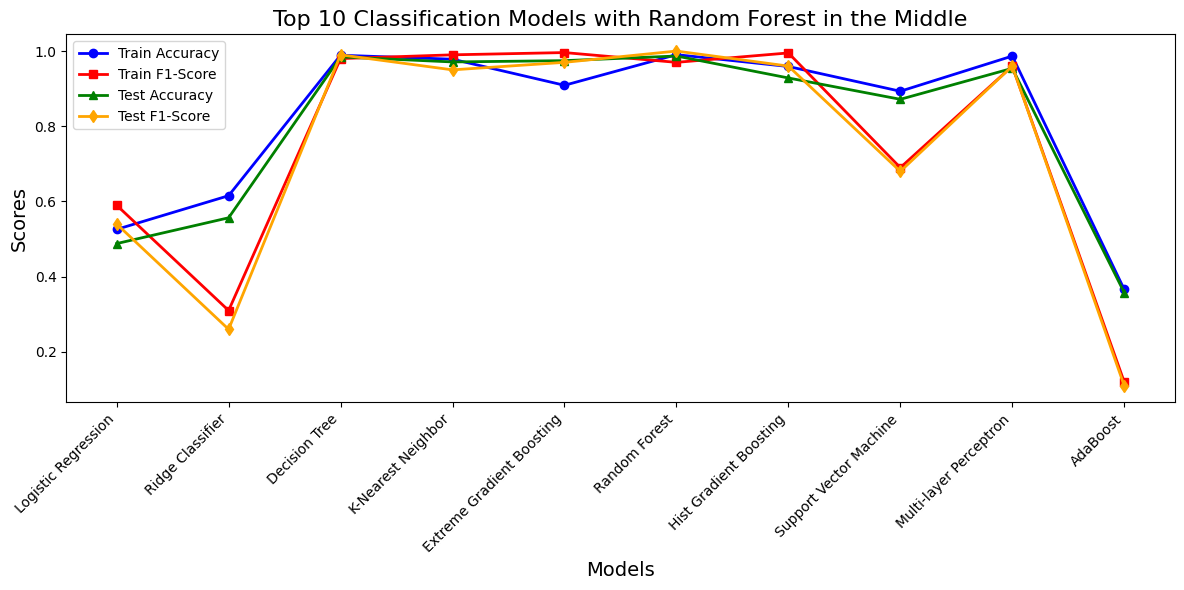} 
  \Description{Bar chart comparing classification model performances, highlighting train and test accuracies alongside F1 scores to evaluate predictive effectiveness.}
  \caption{Comparison of classification model performances, highlighting train and test accuracies alongside F1 scores to evaluate predictive effectiveness.}
  \label{cr}
\end{figure}

In classification tasks for CGPA prediction, 10 models were evaluated, demonstrating varying levels of accuracy and F1-score during training and testing, as illustrated in Figure \ref{cr}. Among them, the Random Forest emerged as the best-performing model, achieving a training accuracy of 99.04\% and a testing accuracy of 98.68\%, alongside a perfect test F1-score of 100\%, underscoring its exceptional generalization capability. Its ensemble learning approach, which integrates multiple decision trees, allows it to efficiently handle complex datasets while maintaining high predictive reliability.   Other tree-based models also displayed competitive performance. The Decision Tree model achieved an impressive training accuracy of 98.89\% and a testing accuracy of 98.45\%, maintaining an F1-score of 99\%. Similarly, the KNN classifier attained a training accuracy of 97.83\% and a test accuracy of 97.11\%, with a slightly lower test F1-score of 95\%, indicating its sensitivity to data distribution.  

Advanced gradient-based models such as XGBoost and Hist Gradient Boosting exhibited strong results, achieving training accuracies of 90.89\% and 95.89\%, with corresponding test accuracies of 97.45\% and 92.84\%, and F1-scores exceeding 96\%, demonstrating their robustness in classification tasks. Likewise, AdaBoost performed moderately well, with a training accuracy of 36.61\% and a test accuracy of 35.77\%, but its F1-score remained relatively low at 11\%, suggesting its limited adaptability to the dataset. Deep learning-based approaches also showcased strong results. The MLP chieved a training accuracy of 98.67\% and a test accuracy of 95.41\%, alongside an F1-score of 96\%, reaffirming the potential of neural networks for such predictive tasks. In contrast, SVM demonstrated a slightly lower performance, with a training accuracy of 89.31\% and a test accuracy of 87.17\%, accompanied by an F1-score of 68\%, reflecting its dependency on feature scaling and kernel selection.  

iConversely, linear models such as Logistic Regression and Ridge Classifier struggled with the dataset, achieving training accuracies of 52.58\% and 61.53\%, while their test accuracies were 48.79\% and 55.66\%, respectively. Their F1-scores remained relatively low (54\% and 26\%), indicating that simpler models may not be suitable for CGPA classification due to the dataset's complexity. Ultimately, Random Forest emerged as the top-performing model, balancing accuracy, interpretability, and computational efficiency. While other models demonstrated strong performance based on their inherent methodologies, Random Forest’s superior generalization ability makes it the most effective choice for CGPA prediction. Figure \ref{cr} visually represents the comparative performance of all evaluated models in terms of accuracy and F1-score across both training and testing phases.  

\subsection{Causal Models}

\begin{figure}[!h]
  \centering
  \includegraphics[width=0.95\textwidth]{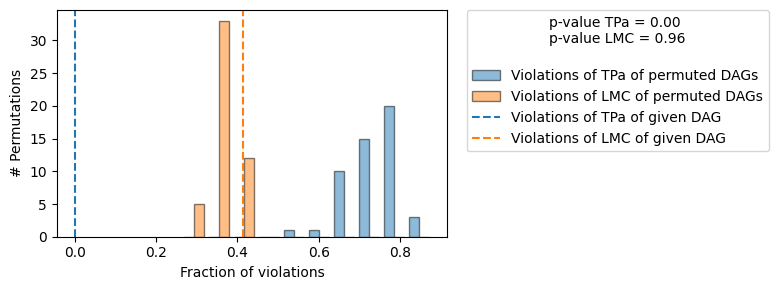} 
  \Description{Bar chart showing evaluation of the causal graph structure with metrics indicating model quality and fit.}
  \caption{Evaluation of the causal graph structure.}
  \label{bar}
\end{figure}

\begin{figure}[!h]
  \centering
  \begin{minipage}{0.48\textwidth}
    \centering
    \includegraphics[width=\textwidth]{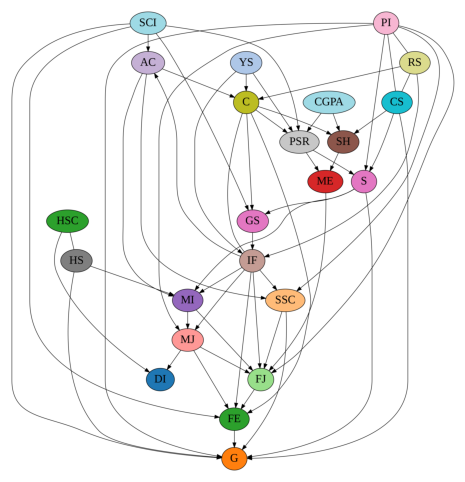}
    \subcaption{Causal model using the PC algorithm}
    \label{causal_pc}
  \end{minipage}
  \hfill
  \begin{minipage}{0.48\textwidth}
    \centering
    \includegraphics[width=\textwidth]{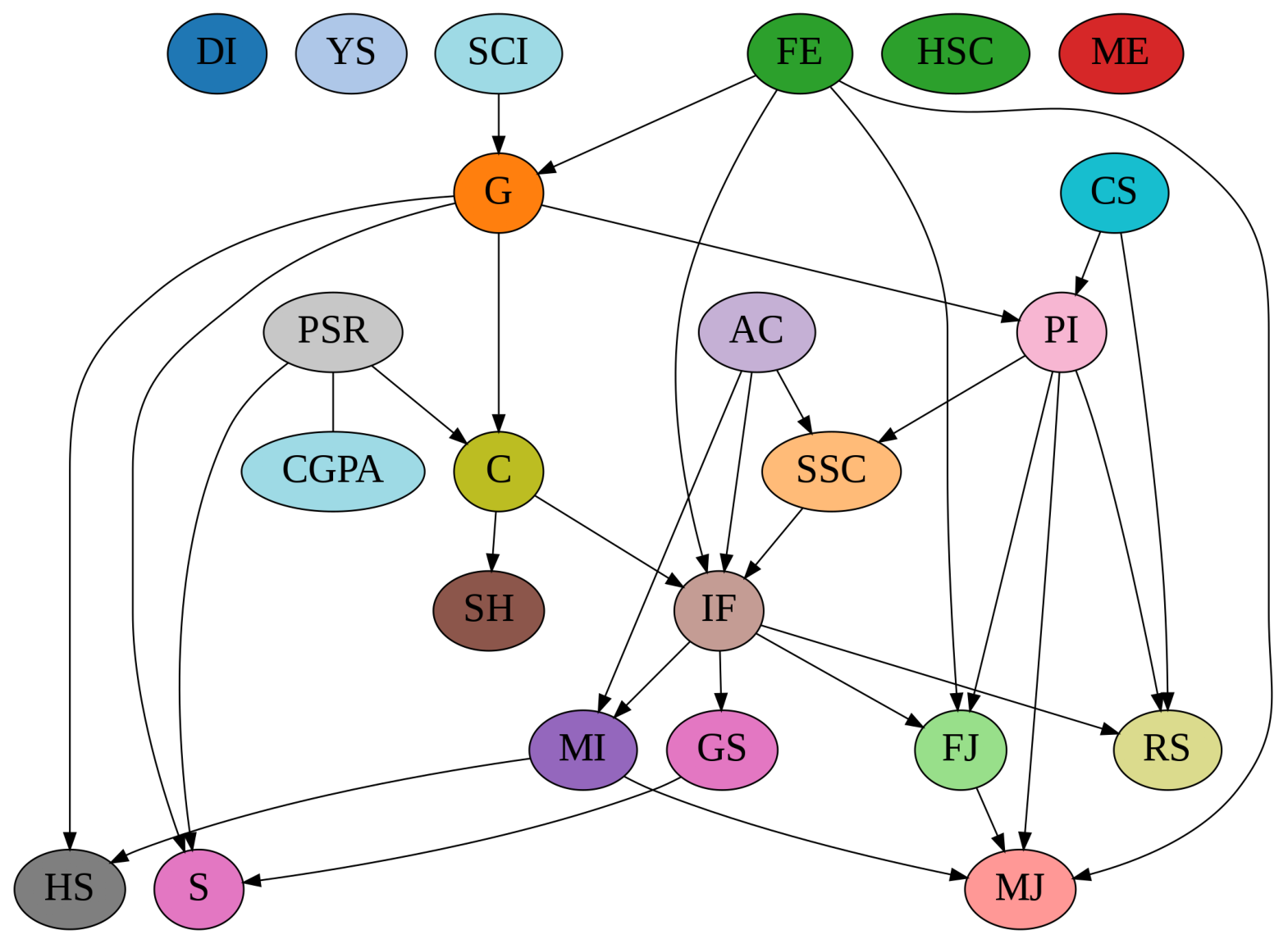}
    \subcaption{Causal model using the GES algorithm}
    \label{causal_ges}
  \end{minipage}
  
  \vspace{0.5\baselineskip}
  
  \begin{minipage}{0.48\textwidth}
    \centering
    \includegraphics[width=\textwidth]{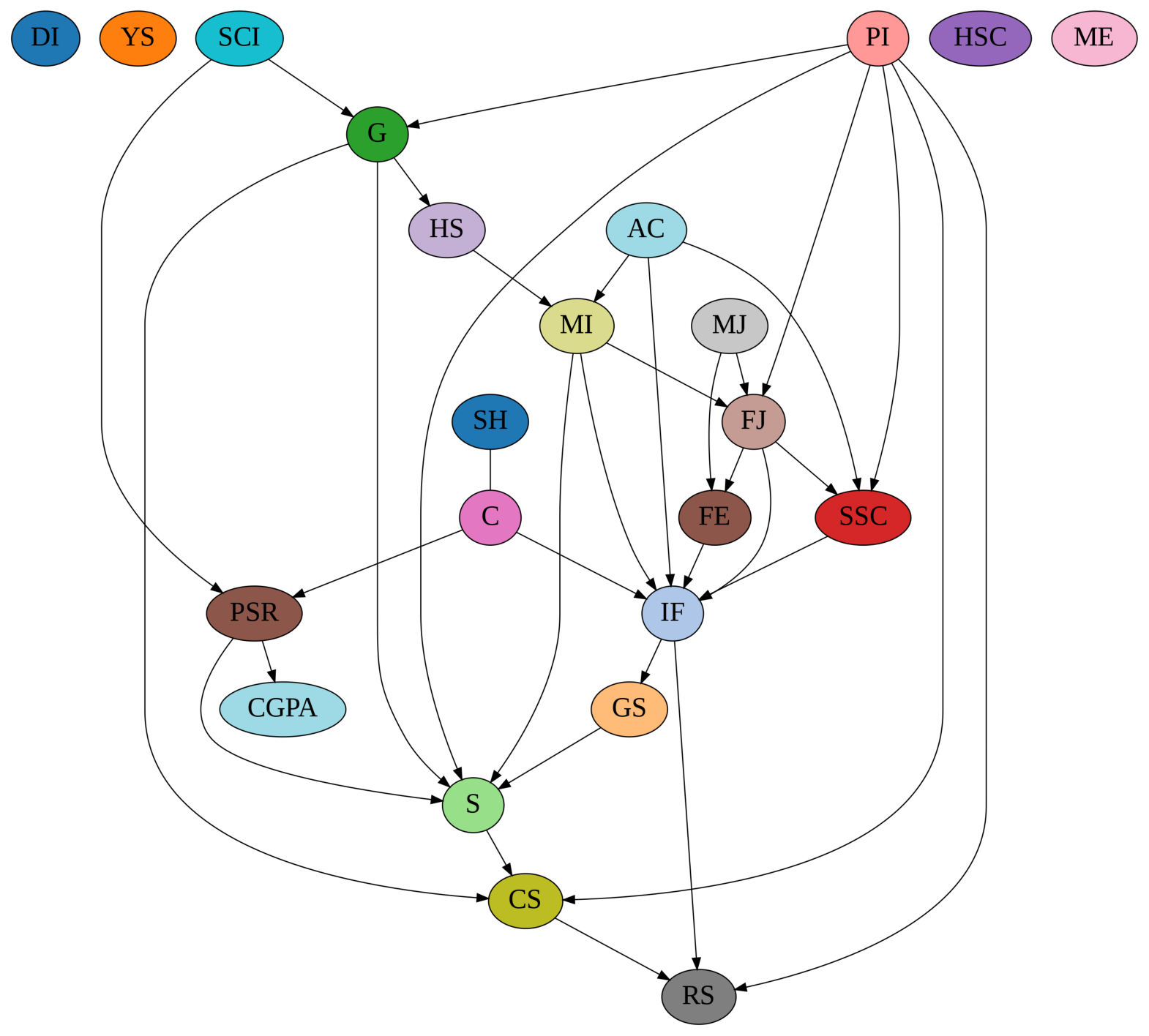}
    \subcaption{Causal model using the Grasp algorithm}
    \label{causal_grasp}
  \end{minipage}
  
  \Description{Three subfigures showing causal models generated using PC, GES, and Grasp algorithms, depicting relationships among variables.}
  \caption{Comparison of causal models generated using PC, GES, and Grasp algorithms.}
  \label{causal_models}
\end{figure}

\begin{figure}[!h]
  \centering
  \includegraphics[width=0.95\textwidth]{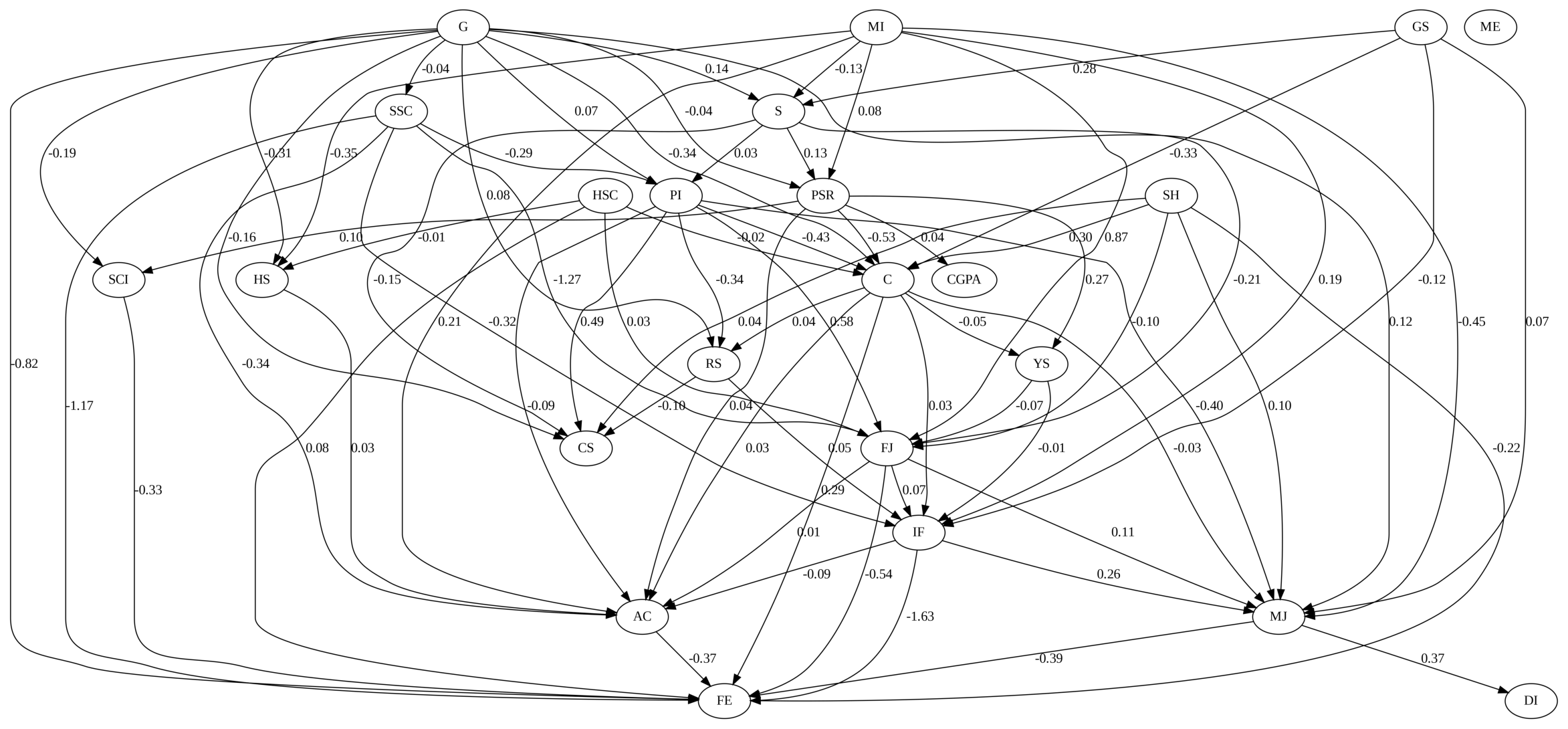} 
  \Description{Causal model generated using the ICA-LiNGAM algorithm displaying weighted edges representing the strength of relationships between features.}
  \caption{Causal model generated using the ICA-LiNGAM algorithm, displaying weighted edges representing the strength of relationships between features.}
  \label{causal_ica}
\end{figure}

First, we analyzed the evaluation results of the hypothesis graph to assess its alignment with the dataset. Subsequently, we examined how different methods—PC, GES, and GRaSP—revealed insights into the variable relationships, contributing to a deeper understanding of the causal structure. The hypothesis graph evaluation results are shown in Figure \ref{bar}. This figure illustrates the fraction of violations for the Triangle Path Assumption and the Latent Markov Condition across permuted Directed Acyclic Graphs. The Triangle Path Assumption violations, with a p-value of zero, suggest strong consistency between the dataset and the proposed causal structure, while the Latent Markov Condition violations, with a p-value of 0.96, indicate areas requiring refinement in indirect relationships.

The PC algorithm revealed intricate relationships where features like FE influenced multiple nodes, including FJ and MJ, while remaining independent of G. SSC played a dual role, influencing IF, PI and AC while maintaining independence from G, Fj, as illustrated in Figure \ref{causal_pc}. SH affected C, CS, and CGPA, remaining independent of ME. Similarly, PSR emerged as a significant feature, impacting C, SCI, and CGPA, while being independent of S. These relationships highlight the centrality of these features in the network.

The GES algorithm identified critical dependencies where FE affected FJ, MJ and IF . SSC emerged as a strong influencer of AC and PI, while SH directly impacted C, as depicted in Figure \ref{causal_ges}. PSR was seen influencing outcomes like SCI and CGPA, emphasizing its importance. These features stood out as pivotal contributors within the causal structure.

The GRaSP method reinforced the significance of FE, SSC, SH, and PSR as impactful features. FE demonstrated widespread influence, directly affecting variables like C,FJ, G,  and MJ, as shown in Figure \ref{causal_grasp}. SSC continued to play a dual role, influencing key outcomes while maintaining independence from others. SH and PSR both highlighted critical connections, with SH influencing C and  PSR significantly impacting CGPA.

The ICALiNGAM analysis provided quantitative insights into feature relationships. FE showed dominant inverse effects on variables such as G, SSC, FJ, MJ, AC, SH, IF, and SCI, with weights ranging from minus 0.32 to minus 1.63. FE also demonstrated positive contributions to HSC and C, with weights of 0.08 and 0.28. SSC exhibited a dual role, negatively influencing AC with a weight of minus 0.34 while positively impacting PI and FJ. SH showed a positive impact on CGPA and C, and PSR revealed significant positive influences on performance and quality-related factors, as illustrated in Figure \ref{causal_ica}, emphasizing their critical role in shaping outcomes.

After analyzing all four causal graphs, we identified FE, SSC, SH, and PSR as the most impactful features. These consistently contributed to key outcomes such as CGPA and demonstrated strong roles in the causal structure. While all features are interconnected, these four stood out as the primary drivers, offering valuable insights for further analysis and decision-making.

\subsection{XAI Insight}

\begin{figure}[!h]
    \centering
    \begin{minipage}{0.99\textwidth}
        \centering
        \includegraphics[width=\textwidth]{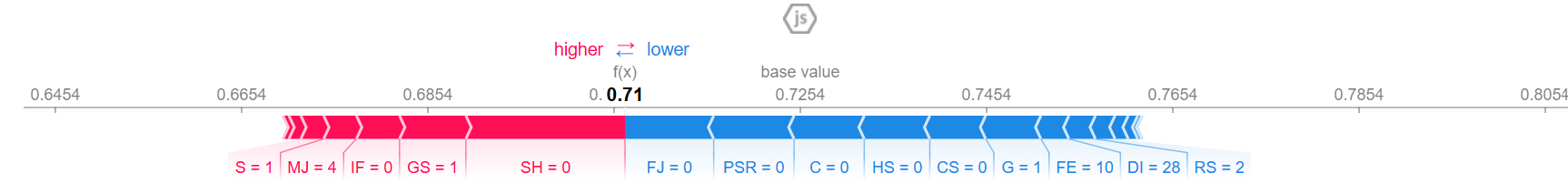}
        \subcaption{Force plot for model interpretation}
        \label{shap}
    \end{minipage}

    \vspace{0.5\baselineskip}

    \begin{minipage}{0.48\textwidth}
        \centering
        \includegraphics[width=\textwidth]{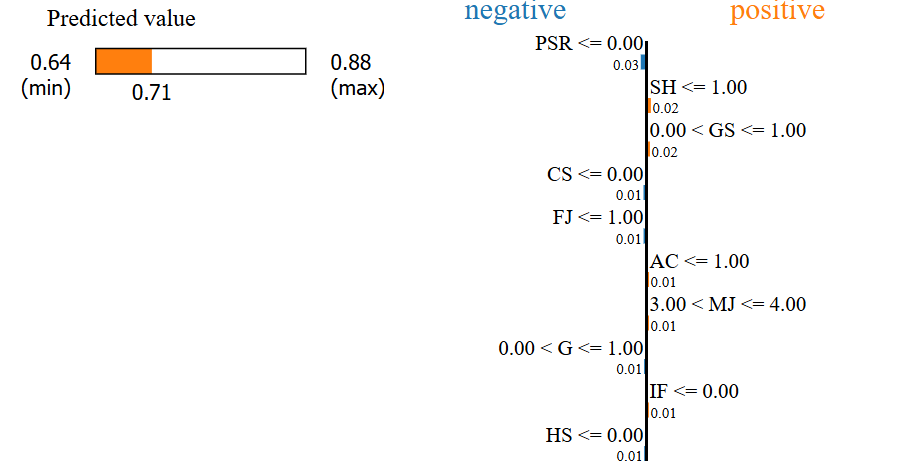}
        \subcaption{LIME interpretation}
        \label{lime}
    \end{minipage}
    \hfill
    \begin{minipage}{0.48\textwidth}
        \centering
        \includegraphics[width=\textwidth]{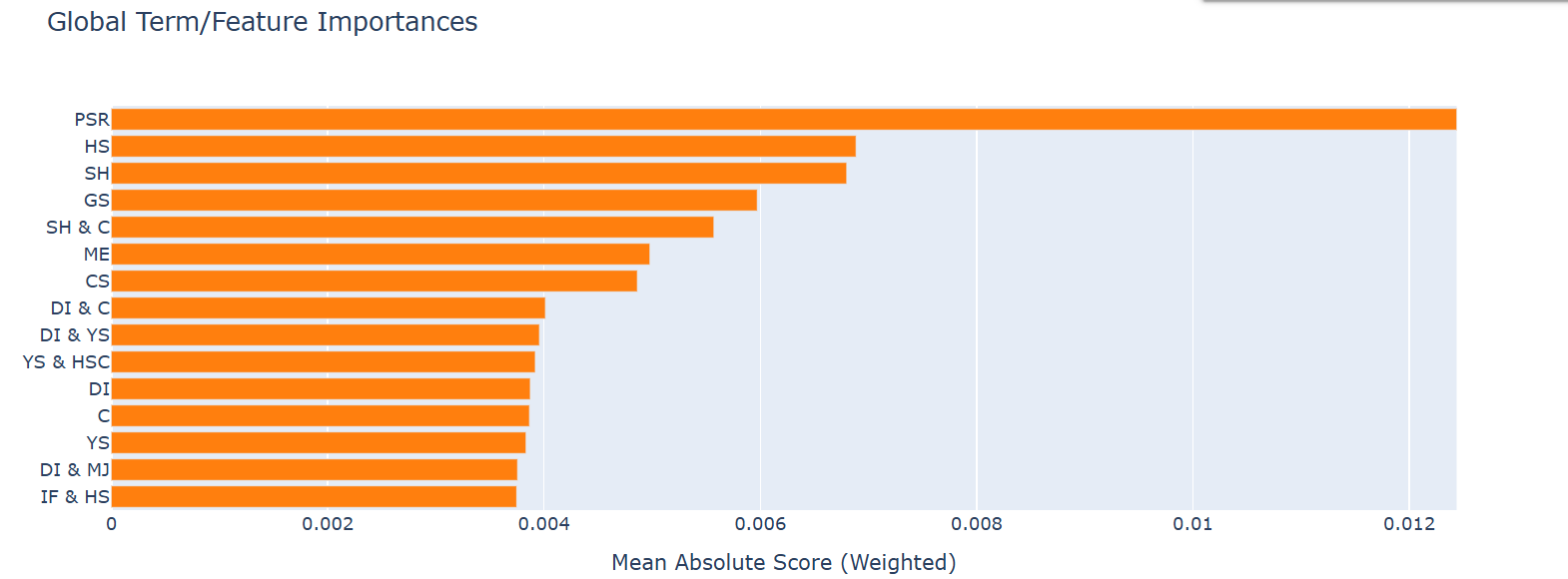}
        \subcaption{Global feature importance}
        \label{in}
    \end{minipage}

    \Description{Model interpretation techniques including a force plot, LIME interpretation, and global feature importance used to explain model behavior and predictions.}
    \caption{Model interpretation techniques: force plot, LIME, and global feature importance for explaining model behavior and predictions.}
    \label{X}
\end{figure}

We found that the best predictive model is Ridge regression. The explainability of the Ridge regression model for CGPA prediction was enhanced through SHAP, LIME, and global feature importance analysis, each contributing distinct insights into the model's behavior.

SHAP analysis showed the base value of the model is 0.7254, representing the average prediction before considering feature contributions. Positive features like S = 1, MJ = 4, and GS = 1 increased the predicted CGPA, as shown in figure \ref{shap}, where red bars indicate positive contributions, while negative features such as FJ = 0, PSR = 0, C = 0, DI = 28, RS = 2, and SH = 0 reduced the predicted value, shown in blue bars. The final predicted CGPA value was 0.71, based on the combined contributions of all features.

LIME analysis indicated that the predicted CGPA ranged from 0.64 to 0.88, with an actual prediction of 0.71, as shown in figure \ref{lime}. Key features like PSR, SH, and GS had the most impact, with PSR and SH contributing negatively, and GS contributing positively. Other contributing features included CS, FJ, AC, MJ, G, IF, and HS, all having varying effects on the predicted CGPA. The specific contributions were observed as PSR less than or equal to 0.00 with -0.03, SH less than or equal to 1.00 with -0.02, GS greater than 0.00 and less than or equal to 1.00 with +0.02, CS less than or equal to 0.00 with +0.01, FJ less than or equal to 1.00 with +0.01, MJ greater than 3.00 and less than or equal to 4.00 with +0.01.

Global Feature Importance analysis confirmed the SHAP and LIME findings, as shown in figure \ref{in}, highlighting PSR, SH, and C as significant contributors. These features were positively correlated with CGPA, likely reflecting the impact of financial stability and academic focus. Other important features included MJ and CS. The combined results from SHAP, LIME, and global feature importance analyses highlighted that features like SH, PSR, C, and Parental Condition were the most influential in predicting CGPA. SH and PSR showed a strong influence, with SH contributing negatively and PSR reducing the predicted CGPA. Parental Condition, such as having a Family Job FJ, was another important factor that lowered the predicted CGPA. These features were identified as key contributors, demonstrating the significant influence of both personal academic habits and external factors on CGPA outcomes.

\subsection{Web App}

\begin{figure}[!h]
    \centering
    \begin{minipage}{0.48\linewidth}
        \centering
        \includegraphics[width=\linewidth,height=4cm]{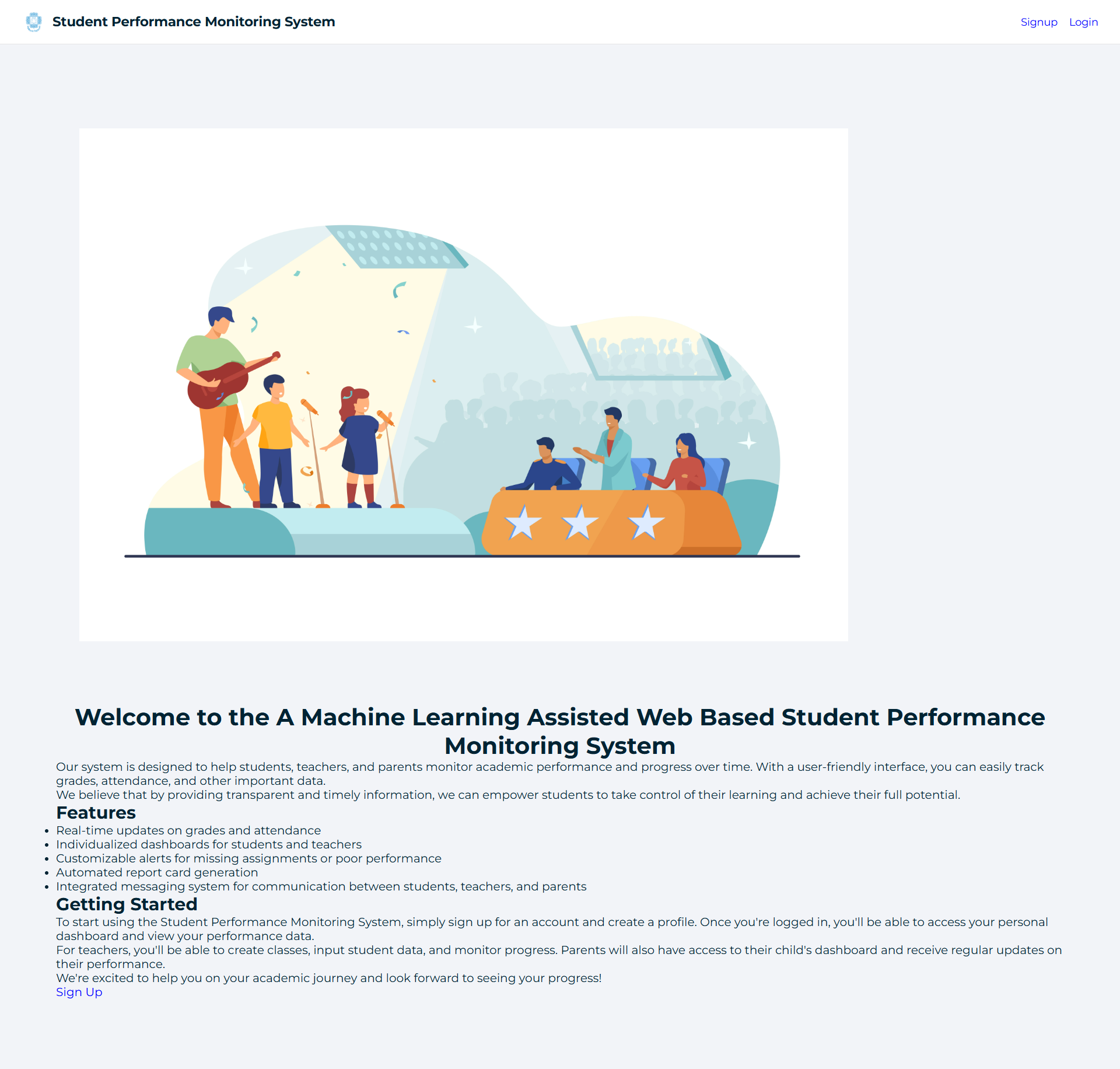}
        \subcaption{The appearance interface where users navigate through the web application.}
        \label{six1}
    \end{minipage}
    \hfill
    \begin{minipage}{0.48\linewidth}
        \centering
        \includegraphics[width=\linewidth,height=4cm]{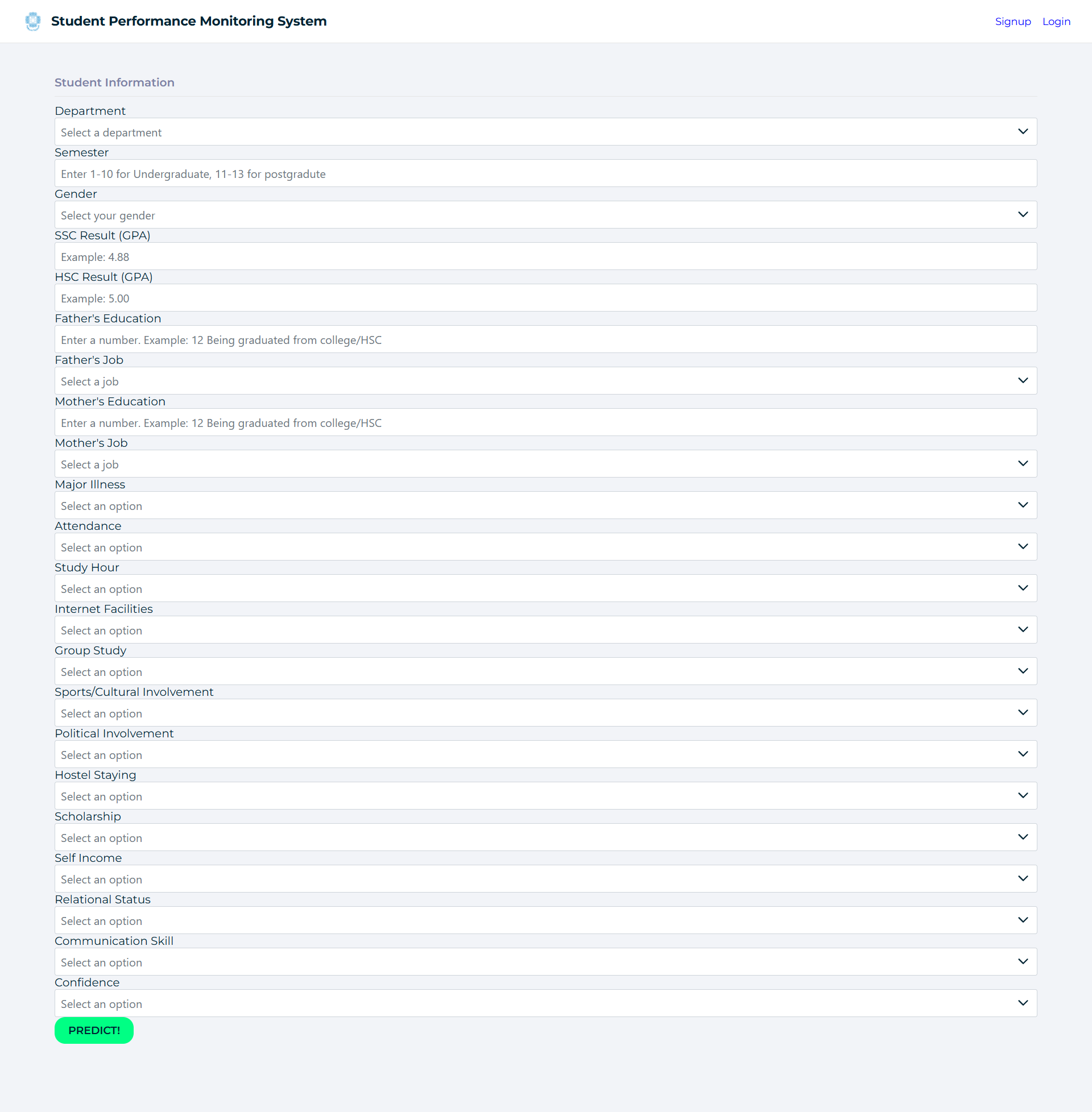}
        \subcaption{The input form where users provide academic and social activity details.}
        \label{six2}
    \end{minipage}

    \vspace{0.5\baselineskip}

    \begin{minipage}{0.48\linewidth}
        \centering
        \includegraphics[width=\linewidth,height=4cm]{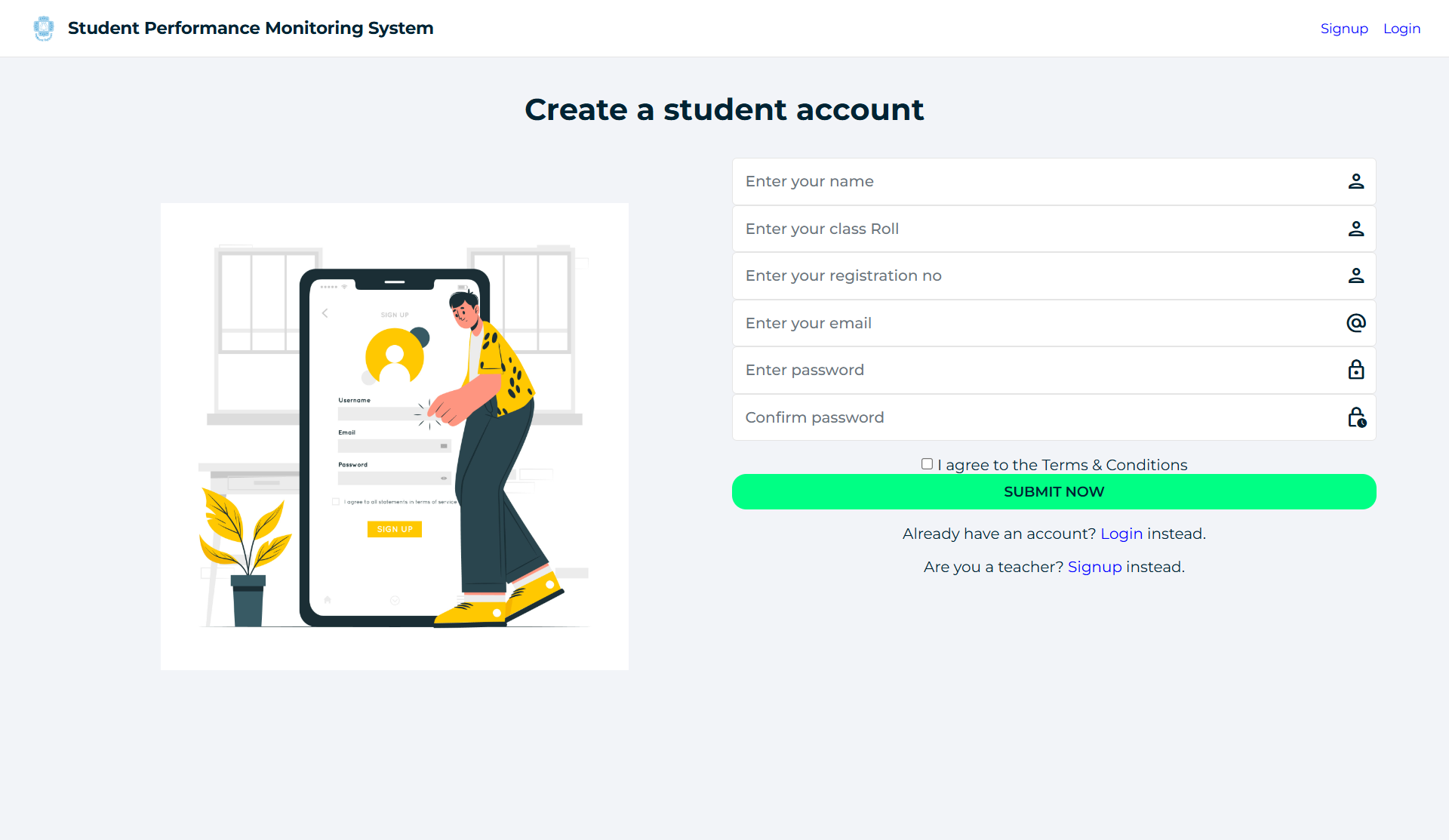}
        \subcaption{The registration form for new users to create an account securely.}
        \label{six3}
    \end{minipage}
    \hfill
    \begin{minipage}{0.48\linewidth}
        \centering
        \includegraphics[width=\linewidth,height=4cm]{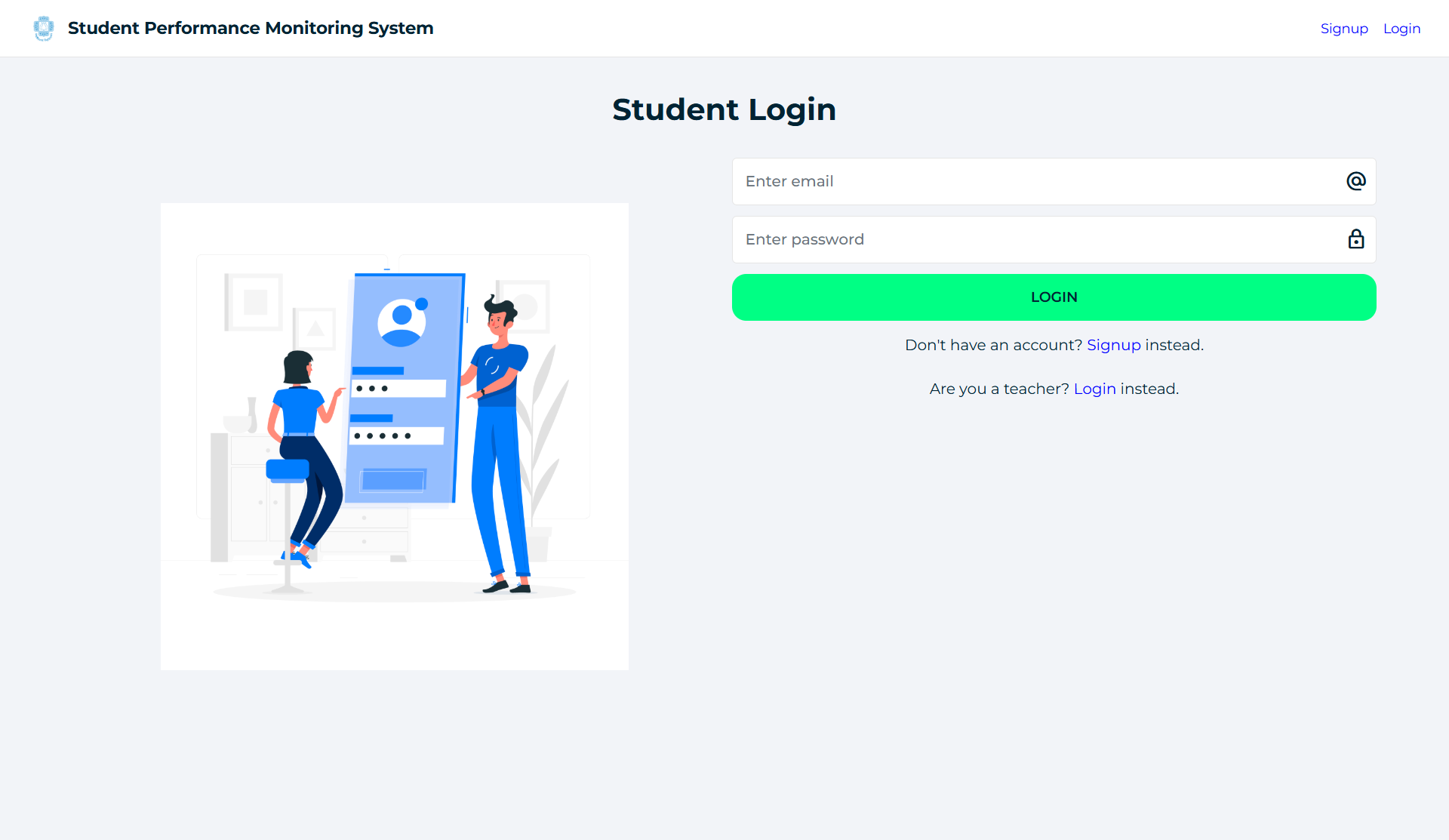}
        \subcaption{The login form where registered users access their accounts.}
        \label{six4}
    \end{minipage}

    \Description{Screen captures from the web application interface, showing user navigation, input form, registration form, and login form.}
    \caption{Screen captures from the web application interface.}
    \label{six}
\end{figure}

In our web application, new users can register securely through the registration form (Fig. \ref{six3}). Once registered, users can log in using their credentials (Fig. \ref{six4}). After logging in, they are presented with an intuitive interface (Fig. \ref{six1}) where they can navigate through the platform and input necessary academic and social activity details (Fig. \ref{six2}). Based on this input, the system predicts their CGPA , helping users make informed decisions about their academic progress and future planning.

The CGPA prediction process considers various factors to ensure accurate and personalized results. Additionally, the web application includes a feedback mechanism, allowing users to share their thoughts and suggestions for platform improvements. This feedback system enables the application to evolve continuously, addressing user needs and enhancing the overall experience. By integrating predictive analytics and user engagement, the web application serves as a valuable tool for students in their academic journey.Our web app can be found in the following GitHub repository: \href{https://github.com/mfarhadhossain/CGPA-Prediction}
{https://github.com/mfarhadhossain/CGPA-Prediction}.

\subsection{Discussion}

The findings highlight significant relationships between student behaviors, demographics, and academic outcomes. For instance, students residing in HS tend to study fewer hours weekly, suggesting potential distractions, while non-HS students study more consistently. GS participation fosters extracurricular involvement, which enhances holistic development. Financial stability emerged as a key factor, with higher-income students exhibiting greater confidence and better performance. Gender differences in class attendance were observed, with female students showing higher attendance rates, potentially impacting academic success positively.
Regression analysis revealed Ridge Regression as the most reliable model for CGPA prediction due to its balance of accuracy and generalization. Classification tasks identified Random Forest as the top performer, excelling in accuracy and F1-score, demonstrating its robustness in handling complex datasets. The causal analysis and XAI techniques both highlighted SH, PSR, FE, and SSC as the key features influencing CGPA. While causal analysis provided insights into the relationships between these features and other variables, XAI techniques confirmed their significance in CGPA prediction, with SH and PSR being the most influential. The consistency between these approaches strengthens the validity of selecting these features for CGPA prediction. Thus, SH, PSR, FE, and SSC are identified as the primary drivers, offering valuable insights for improving student performance and academic strategies.

\section{Conclusions}

This study provides valuable insights into the factors influencing student academic outcomes and the effectiveness of predictive and interpretative tools in an educational context. Ridge Regression was identified as the most reliable model for CGPA prediction, balancing accuracy and generalization. Random Forest excelled in classification tasks, demonstrating robustness in handling complex datasets. Causal analysis highlighted the critical roles of key features, including FE, SSC, SH, and PSR, which consistently influenced academic performance metrics like CGPA.

The integration of XAI techniques, such as SHAP and LIME, significantly enhanced the interpretability of the predictive models, offering a detailed understanding of feature contributions. By aligning insights from causal analysis and XAI methods, features like SH, PSR, and SSC were validated as pivotal contributors to CGPA predictions. This alignment not only strengthens the reliability of the findings but also provides actionable recommendations for targeted interventions. Ultimately, this work underscores the importance of combining predictive modeling with interpretative techniques to develop transparent, reliable, and actionable strategies for improving student outcomes, aiding educators and administrators in fostering academic success.

In addition, a key contribution of this study is the development of a web application designed to facilitate the use of the prediction model and the interpretative insights. The application provides a user-friendly interface for students and educators to easily analyze and act on CGPA-related insights. By offering features such as predictive modeling and real-time interpretation, the web application empowers both students and educators to take proactive steps to improve academic performance.

However, the study is not without limitations. The dataset's scope was constrained to specific variables and contexts, potentially limiting the generalizability of findings across broader populations. Additionally, external factors such as emotional well-being or unforeseen disruptions were not included in the analysis, which may impact academic outcomes. Future work can address these limitations by incorporating a more diverse dataset and exploring additional variables, such as behavioral or socio-emotional factors. Furthermore, extending the approach to include longitudinal analysis could offer deeper insights into the evolving impact of critical features on academic success. The integration of more advanced machine learning models and real-time feedback mechanisms, along with the web application’s continuous updates, may also enhance prediction accuracy and practical applicability, creating a more comprehensive framework for fostering student success.

\bibliographystyle{acmtrans}
\bibliography{ref}

\end{document}